\documentclass{article}

 \usepackage[eandd, final]{neurips_2026}

\usepackage[utf8]{inputenc} 
\usepackage[T1]{fontenc}    
\usepackage{hyperref}       
\usepackage{url}            
\usepackage{booktabs}       
\usepackage{amsfonts}       
\usepackage{nicefrac}       
\usepackage{microtype}      
\usepackage{xcolor}         

\usepackage{amsmath, amssymb, amsthm}
\usepackage{graphicx}
\usepackage{diagbox}
\usepackage{hyperref}
\usepackage{multirow}
\usepackage{makecell}
\usepackage{tabularx}
\usepackage{algorithm}
\usepackage{algpseudocode}
\usepackage{amsmath}
\usepackage{caption}

\usepackage{bbding}
\usepackage{array}
\usepackage{booktabs}
\usepackage{hyperref}

\usepackage{algorithm}
\usepackage{booktabs}
\usepackage{multirow}
\usepackage{amsmath}
\usepackage{amssymb}
\usepackage{colortbl}
\usepackage{xcolor}
\usepackage{graphicx}  
\usepackage{float}  
\usepackage[accsupp]{axessibility}  
\usepackage{wrapfig}
\usepackage{tablefootnote}
\usepackage{titletoc}

\newcommand{\eg}{\textit{e.g. }}
\newcommand{\ie}{\textit{i.e. }}
\newcommand{\etc}{\textit{etc. }}
\definecolor{heading}{HTML}{FB08DE}
\definecolor{reading}{HTML}{FF0808}
\definecolor{truncate}{HTML}{2A1AF2}
\definecolor{figure}{HTML}{00AA00}

\title{MPDocBench-Parse: Benchmarking Practical Multi-page Document Parsing
}

\author{
Bangbang Zhou\textsuperscript{1}\thanks{Interns at Tongyi Lab, Alibaba Group},
Hangdi Xing\textsuperscript{2},
Yifan Chen\textsuperscript{1},
Jianjun Xu\textsuperscript{1},
Qi Zheng\textsuperscript{2}\thanks{Project leader},
Feiyu Gao\textsuperscript{2}, \\
\textbf{Zhibo Yang\textsuperscript{2}},
\textbf{Shuai Bai\textsuperscript{2}},
\textbf{Ming Yan\textsuperscript{2}},
\textbf{Jieping Ye\textsuperscript{2}},
\textbf{Hongtao Xie\textsuperscript{1}}\thanks{Corresponding author}
\\[0.6ex]
\textsuperscript{1~}University of Science and Technology of China\quad
\textsuperscript{2~}Tongyi Lab, Alibaba Group\quad \\ [1ex]
Project Page: \href{https://github.com/Tongyi-Zhiwen/Qwen-Doc}{https://github.com/Tongyi-Zhiwen/Qwen-Doc} \\
}

\begin{document}

\maketitle

\begin{abstract}
Document parsing converts visually rich documents into machine-readable structured representations, forming a crucial foundation for information systems. Although many benchmarks have been proposed for document parsing, they remain inadequate for realistic scenarios. Existing benchmarks either focus on specific tasks or assess only single-page, text-centric settings, making them insufficient for practical multi-page parsing. Moreover, they lack fine-grained evaluation of semantic continuity, hierarchical structure recovery, and visual content preservation. To address these gaps, we propose \texttt{MPDocBench-Parse}, a benchmark for multi-page document parsing in real-world applications. It contains 433 manually annotated documents with 3,246 pages, covering 15 document types in English and Chinese, with diverse layout styles, and supports document-level end-to-end evaluation. We further design a comprehensive protocol for content fidelity and logical structure, covering text, table, and formula recognition, truncated text and table merging, figure extraction, reading order, and heading hierarchy recovery. Experiments show that, while existing models perform well on basic text extraction, they still suffer clear limitations in semantic continuity integration, visual content parsing, and hierarchical structure recovery. MPDocBench-Parse provides a unified foundation for advancing document parsing toward more realistic scenarios.
\end{abstract}

\begin{figure}[h]
\centering
    \includegraphics[width=1.0\linewidth]{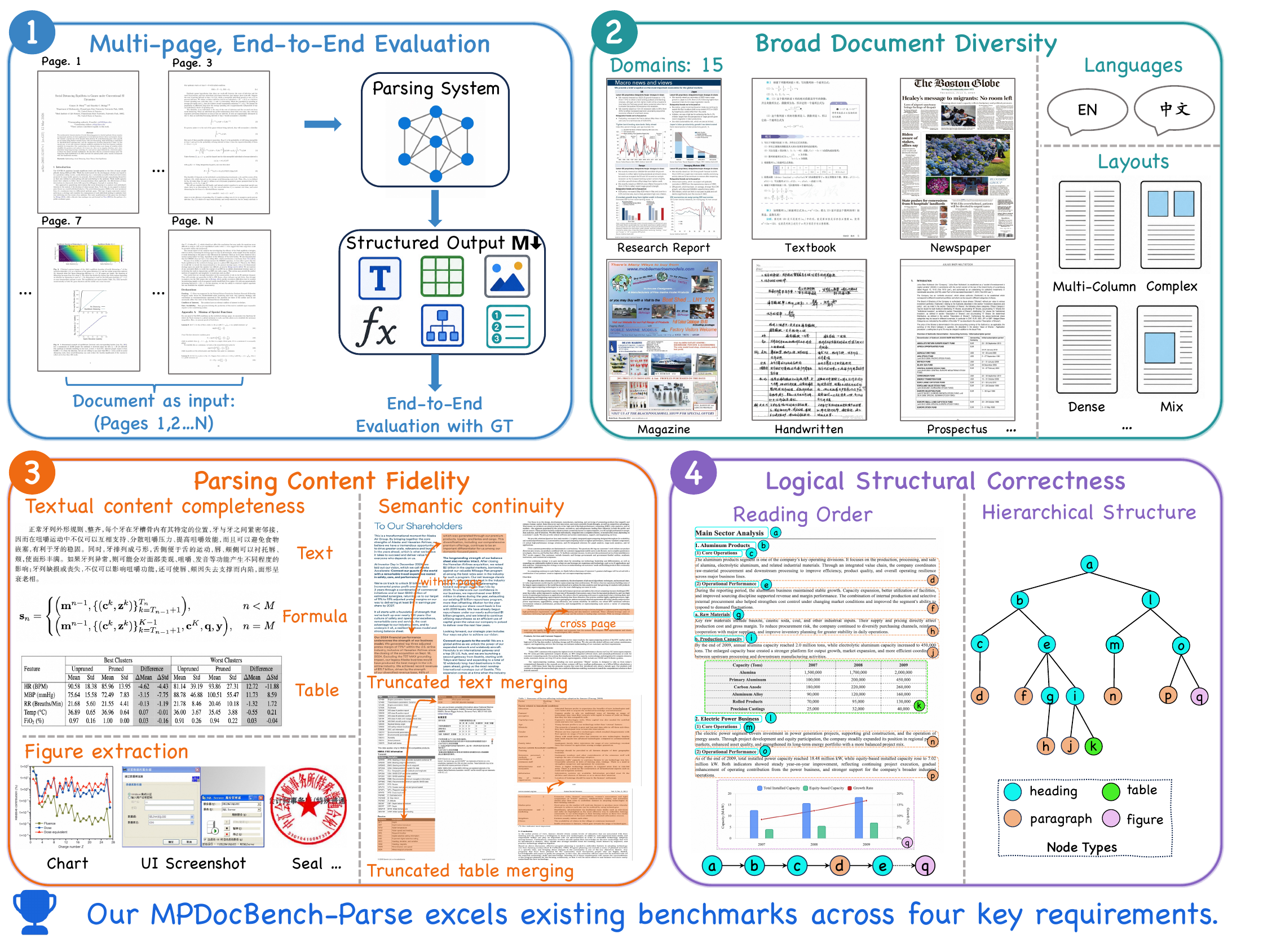}
    \caption{Overview of MPDocBench-Parse. Compared with existing benchmarks, MPDocBench-Parse better supports practical document parsing through four key requirements: multi-page, end-to-end evaluation, broad document diversity, parsing content fidelity, and logical structural correctness.}
    \label{fig:motivation}
    \vspace{-20pt}
\end{figure}

\section{Introduction}
Documents are a fundamental medium for storing and communicating information across real-world domains such as economics, law, scientific research, \etc In the era of AI, converting document images into machine-readable structured formats, such as Markdown or JSON, has become a critical capability. Such structured document parsing not only provides high-quality corpora for training Large Language Models (LLMs) and Multimodal Large Language Models (MLLMs), but also serves as an essential external knowledge interface for Retrieval-Augmented Generation (RAG) systems and AI agents. 
As document parsing is increasingly expected to support real-world deployment in enterprise and industrial scenarios~\cite{zhang2026parsebench}, the parsing benchmarks should evaluate models under settings that closely reflect practical usage conditions.

However, existing benchmarks remain insufficient in this regard. Early task-specific benchmarks~\cite{PubLayNet,Tablebank,Im2Latex-100K,layoutreader,RobustReading,hrdoc} decompose the document parsing task into multiple isolated subproblems, \ie{layout analysis}, text recognition, formula recognition, and heading recovery. For example, PubLayNet~\cite{PubLayNet} and DocLayNet~\cite{DocLayNet} focus on page-level layout analysis, while PubTabNet~\cite{PubTabNet} and SciTSR~\cite{scitsr} evaluate table recognition performance on cropped table images. 
They abstract away from the actual document parsing pipeline in practical scenarios, where systems must process complete documents and produce directly usable structured outputs.
Recent end-to-end benchmarks~\cite{fox,olmocr,omnidocbench,got2,Nougat,readoc,ccocr,docptbench,mdpbench} move closer to practical usage by evaluating structured outputs such as Markdown or JSON directly from document images. Nevertheless, they remain insufficient for realistic deployment in several respects. First, most benchmarks are designed for single-page parsing~\cite{omnidocbench,logicsparsing,mdpbench}, whereas practical document parsing often involves multi-page inputs, making cross-page semantic continuity and document-level organization difficult to assess. Second, existing evaluation protocols~\cite{ccocr,olmocr,docptbench} largely focus on textual correctness, such as text, table, and formula recognition, while providing limited assessment of beyond-text capabilities, particularly document hierarchy recovery and multimodal visual content. 
Third, many benchmarks~\cite{Nougat,readoc} offer limited coverage of document scenarios commonly encountered in enterprise and industrial settings, due in part to restricted domain diversity and an over-reliance on academic-style documents.

As illustrated in Fig.~\ref{fig:motivation}, the limitations of existing benchmarks motivate us to argue that a practical parsing benchmark should satisfy four key requirements:
\textbf{(1)} \emph{multi-page, end-to-end evaluation}, so that models are assessed under realistic parsing scenarios rather than isolated task-level or page-level settings; \textbf{(2)} \emph{broad document diversity}, so that models can be reliably evaluated across domains, languages, and layout styles; \textbf{(3)} \emph{parsing content fidelity}, so that the parsed output preserves not only textual content completeness but also semantic continuity across layout fragmentation and page breaks, as well as multimodal visual information; and \textbf{(4)} \emph{logical structural correctness}, so that the output retains both reading order and hierarchical relationships for proper document organization.

To this end, we introduce \textbf{MPDocBench-Parse}, a benchmark for multi-page document parsing designed to better reflect realistic document processing scenarios. As shown in Fig.~\ref{fig:motivation} and Tab.~\ref{tab:benchmark_domain_distirbution}, MPDocBench-Parse contains {15} document types in both English and Chinese. This diversity enables a more robust evaluation under realistic variations in domain, language, and layout. Furthermore, its fine-grained manual annotations support not only end-to-end evaluation but also single-task evaluation of individual parsing capabilities across different subtasks.

To assess parsing content fidelity, MPDocBench-Parse evaluates recognition quality on text, tables, and formulas as a measure of textual content completeness, and further introduces truncated text and table merging to test whether semantically continuous content can be correctly merged when paragraphs or tables are split by complex layout boundaries or page breaks. In addition, MPDocBench-Parse is the first benchmark to evaluate figure extraction, addressing a longstanding gap in current document parsing evaluation for visually grounded content such as charts, seals, flowcharts, and product images, which are essential for understanding visually rich documents. 
As for logical structural correctness, MPDocBench-Parse evaluates both reading order and heading hierarchy, requiring models to recover not only local reading sequences but also the global hierarchical organization of documents. Such structure-aware parsing is critical for downstream applications such as section-aware RAG~\cite{bookrag,yeh2026kohakurag}, structured summarization, and fine-grained question answering~\cite{deepread,multidocfusion}.

\begin{table*}[t]
\caption{
Comparison with existing benchmarks on document diversity, evaluation coverage, and paradigm, highlighting that our MPDocBench-Parse better matches practical parsing scenarios.
}\label{tab:benchmark_domain_distirbution}
\centering
\renewcommand{\arraystretch}{1.0}
\resizebox{1.0\textwidth}{!}{
\begin{tabular}{l|c|ccccccccc|cc}
\toprule
\textbf{\multirow{2}[4]{*}{Benchmark}} & \multirow{2}[4]{*}{\makecell[c]{\textbf{Domain}\\\textbf{Diversity}}} & \multicolumn{9}{c|}{\textbf{Evaluation Coverage}} & \multicolumn{2}{c}{\textbf{Evaluation Paradigm}} \\
\cmidrule{3-13}
& & \makecell{Layout\\Analysis}& \makecell{Text\\Recognition} & \makecell{Table\\Recognition} & \makecell{Formula\\Recognition} & 
\makecell{Reading\\Order} & \makecell{Heading\\Hierarchy} & \makecell{Truncated\\Text Merging} & \makecell{Truncated\\Table Merging} & \makecell{Figure\\Extraction} & \makecell{Single\\Task} & \makecell{End-to-End} \\
\midrule
\multicolumn{13}{l}{\textbf{\textit{Single-Task Benchmarks}}} \\
\midrule
\makecell[l]{PubLayNet~\cite{PubLayNet}, DocLayNet~\cite{DocLayNet}} & \makecell[c]{1,5} & \CheckmarkBold & & & & & & & & &  \CheckmarkBold &  \\
\midrule
Robust Reading~\cite{RobustReading} & 1 &  & \CheckmarkBold & & & & & & & & \CheckmarkBold &   \\
\midrule
\makecell[l]{PubTabNet~\cite{PubTabNet},SciTSR~\cite{scitsr}} & \makecell{1,1} &  & & \CheckmarkBold & &  & & &  & &  \CheckmarkBold &  \\
\midrule
Im2Latex-100K~\cite{Im2Latex-100K},UniMER-Test~\cite{wang2024unimernetuniversalnetworkrealworld} & 1,1 &  &  & & \CheckmarkBold &  &  &  & & & \CheckmarkBold & \\ 
\midrule
Layoutreader~\cite{layoutreader} & 1 &  &  & &  & \CheckmarkBold &  &  &  & & \CheckmarkBold &  \\
\midrule
HRDoc~\cite{hrdoc}, DocHieNet~\cite{dochienet} & 1, 5 & & &  &  &  & \CheckmarkBold &  & &  &  \CheckmarkBold &  \\
\midrule
\multicolumn{13}{l}{\textbf{\textit{Single-Page Benchmarks}}} \\
\midrule
Fox~\cite{fox} & 2 & & \CheckmarkBold &  &  & &  &  & &  &  &  \CheckmarkBold \\

Nougat~\cite{Nougat}, GOT OCR 2.0~\cite{got2} & 1, 2 &  & \CheckmarkBold & \CheckmarkBold & \CheckmarkBold &  &  &  & &  &  &  \CheckmarkBold  \\

olmOCR-Bench~\cite{olmocr} & 7 &  & \CheckmarkBold & \CheckmarkBold & \CheckmarkBold & \CheckmarkBold  &  & & & &  &  \CheckmarkBold   \\

OmniDocBench~\cite{omnidocbench} & 9 & \CheckmarkBold & \CheckmarkBold & \CheckmarkBold & \CheckmarkBold & \CheckmarkBold &  &  &  &  &  \CheckmarkBold &  \CheckmarkBold   \\

\midrule
\multicolumn{13}{l}{\textbf{\textit{Multi-Page Benchmarks}}} \\
\midrule
READoc~\cite{readoc} & 3 & & \CheckmarkBold & \CheckmarkBold & \CheckmarkBold & \CheckmarkBold & \CheckmarkBold &  & &  &  &  \CheckmarkBold \\
\midrule
\rowcolor{gray!15} \textbf{MPDocBench-Parse} & 15 & \CheckmarkBold & \CheckmarkBold & \CheckmarkBold & \CheckmarkBold & \CheckmarkBold & \CheckmarkBold & \CheckmarkBold & \CheckmarkBold & \CheckmarkBold & \CheckmarkBold & \CheckmarkBold   \\
\bottomrule
\end{tabular}
}
\end{table*}

Based on MPDocBench-Parse, we conduct a comprehensive evaluation of existing specialized and general VLMs. Results in Tab.~\ref{tab:main_results} lead to the following conclusions: 
(1) Current models perform reasonably well on relatively straightforward textual content extraction tasks, such as text, table, and formula recognition, but suffer substantial performance degradation on more challenging semantic continuity and visual localization tasks, such as truncated text and table merging, and figure extraction.
(2) Heading hierarchy prediction remains difficult for all models, although end-to-end parsing models~\cite{fireredocr,deepseekocr2,qianfanocr,chandraocr2} generally outperform pipeline-based systems~\cite{mineru25,glmocr,monkeyocr}.
(3) More importantly, MPDocBench-Parse reveals limitations that are largely hidden by existing benchmarks~\cite{omnidocbench}: strong performance on isolated page-level does not necessarily translate to robust parsing in realistic multi-page documents.
We hope these findings will encourage future research to place greater emphasis on these challenges in document parsing. Our main contributions are summarized as follows:
\begin{itemize}
    \item We introduce MPDocBench-Parse, an end-to-end benchmark for practical multi-page document parsing. It consists of {433} manually annotated PDF documents, comprising {3,246} pages, spanning 15 diverse document types in both English and Chinese.
    \item We propose a comprehensive evaluation protocol for multi-page document parsing that jointly measures parsing content fidelity and logical structural correctness, covering content completeness, semantic continuity, multimodal preservation, and structural organization.
    \item We benchmark representative specialized and general vision-language models under the proposed protocol, providing a systematic analysis of their strengths, limitations, and failure modes in realistic multi-page document parsing.
\end{itemize}

\section{Related Work}
\textbf{Methods for Document Parsing.}
Existing document parsing methods can be broadly grouped into pipeline-based tools, general vision-language models (VLMs), and specialized VLM parsers. Pipeline-based methods~\cite{mineru,paddleocrv3,docling,layoutparser} decompose parsing into subtasks such as layout detection, OCR, element recognition, and reading order reconstruction. They are efficient and controllable, but their separately designed parsing modules often suffer from error accumulation. General VLMs~\cite{gemini31,chatgpt,claude,bailian,qwen3vl} offer a simple end-to-end interface by directly generating structured outputs such as Markdown or HTML from document images, but they often struggle with dense text, fine-grained visual grounding, and hallucinations. 
Recent specialized VLM parsers~\cite{olmocr,mineru25,monkeyocr,dolphinv2,paddleocrvl} attempt to address document parsing with unified VLM-style modeling by training multiple parsing tasks within a single model. For example, MonkeyOCR~\cite{monkeyocr} decouples document parsing into structure detection, content recognition, and relation prediction, where the content recognition module can jointly handle text, tables, and formulas. MinerU2.5~\cite{mineru25} goes a step further by integrating these components into a single model that first performs layout analysis and then recognizes the parsed content regions.

\textbf{Benchmarks for Document Parsing.}
Early benchmarks focus on single, isolated tasks in document parsing, including layout analysis~\cite{PubLayNet,DocLayNet,M6Doc,da2023vision}, OCR~\cite{RobustReading}, and table structure recognition~\cite{PubTabNet,fintabnet}. Although these benchmarks are useful for measuring specific abilities, they fail to evaluate the overall capability of models to parse complete documents into structured outputs.
Recent benchmarks have shifted toward end-to-end document parsing. OmniDocBench~\cite{omnidocbench} introduces a multi-dimensional evaluation framework covering full-page assessment across nine document types. READoc~\cite{readoc} further extends evaluation to PDF-to-Markdown extraction and introduces heading hierarchy assessment. DocPTBench~\cite{docptbench} and MDPBench~\cite{mdpbench} improve realism by considering photographed documents and multilingual scenarios. However, existing benchmarks are still limited in document diversity and evaluation coverage. Many of them are biased toward academic or web-derived documents, and they provide insufficient evaluation of multi-page challenges, such as cross-page table merging, truncated text recovery, visual content preservation, and document-level logical structure. The lack of a practical multi-page document parsing benchmark leaves a significant gap in evaluating content faithfulness, multimodal information preservation, and structural correctness in realistic scenarios.

\begin{figure}[t]
    \centering
    \vspace{-20pt}
    \includegraphics[width=1.0\linewidth]{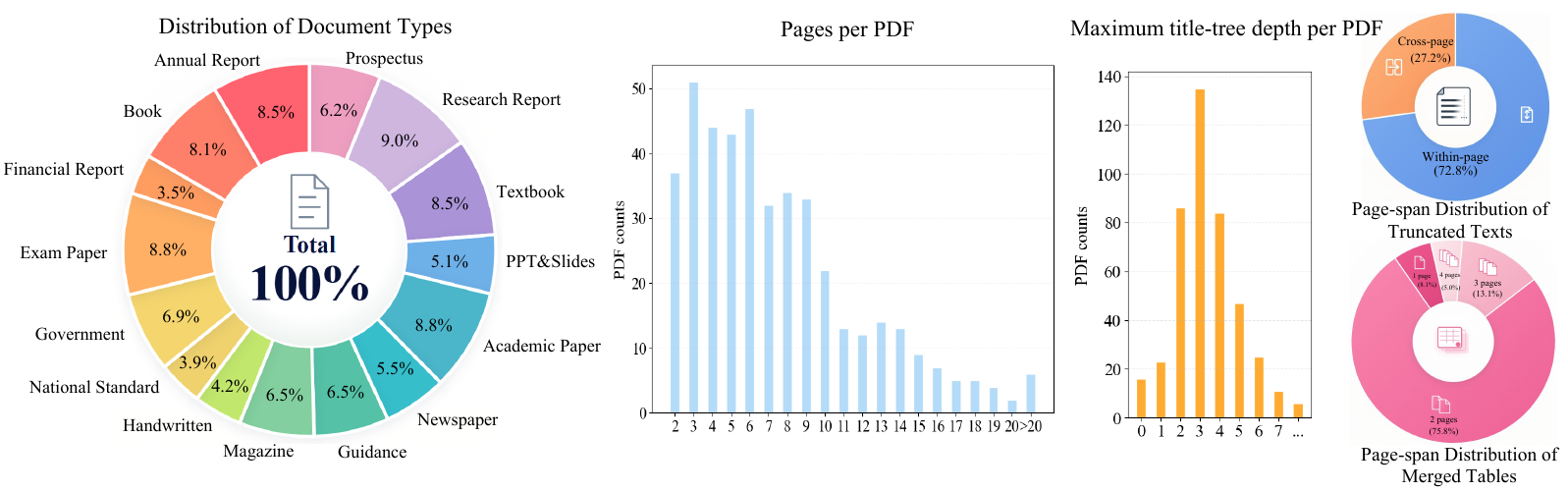}
    \vspace{-20pt}
   \caption{Dataset statistics of MPDocBench-Parse.}\label{fig:data_statistics}
    \vspace{-12pt}
\end{figure}

\section{MPDocBench-Parse}

\subsection{Multi-page Document Data Collection}
To ensure data diversity, we systematically review the document types covered by recent document parsing benchmarks~\cite{omnidocbench,dochienet,logicsparsing,logicsparsingv2,monkeyocr} and document question answering benchmarks~\cite{mmlongbenchdoc,longdocurl,unidoc,docbench}. After aggregation and filtering, we identify 15 document types that cover a wide range of real-world scenarios, as illustrated in Fig.~\ref{fig:data_statistics}.
Using these document types as keywords, we collect about 5,000 raw PDF documents from the Internet through manual downloading. 
To achieve sufficient diversity in layout structures, heading hierarchies, and visual content, we further employ external annotators to manually screen the collected documents, prioritizing those with complex layouts and rich content.

Meanwhile, we balance the data across document types to avoid the over- or under-representation of any category. In addition, we include a number of long documents to further increase the benchmark difficulty and better evaluate model performance in multi-page document parsing scenarios.
After the above screening and balancing process, we finally obtain {433} multi-page documents, comprising {3246} page images (see Fig.~\ref{fig:data_statistics} for the proportion of each document type and  document length distribution). Overall, MPDocBench-Parse covers both relatively simple documents and challenging ones with complex hierarchical structures and rich visual elements, enabling a more comprehensive and practical evaluation of document parsing systems.

\begin{figure}[t]
\includegraphics[width=1.0\linewidth]{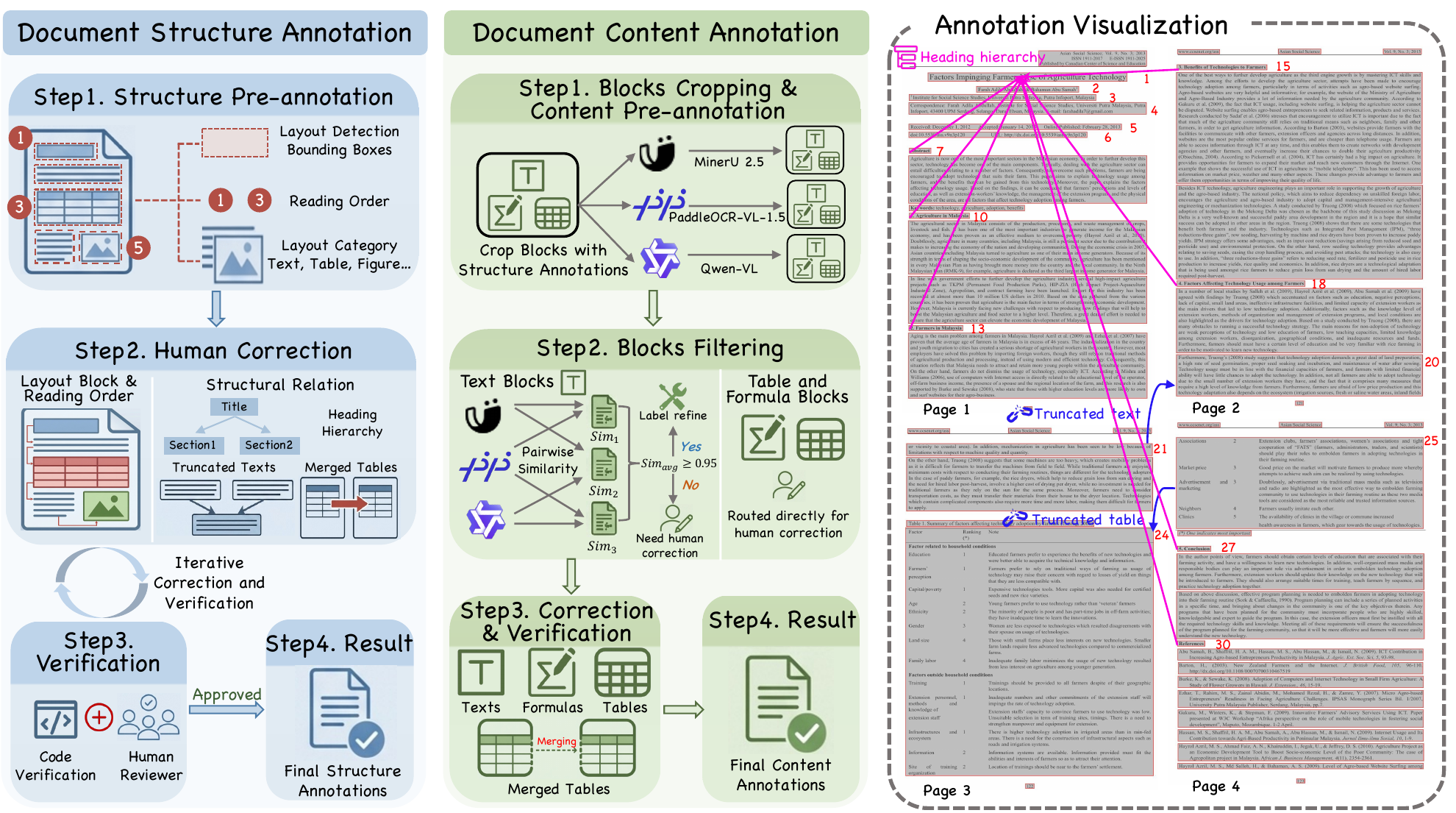}
\caption{MPDocBench-Parse's construction pipeline. The left and middle panels illustrate the structure and content annotation process. The right panel shows annotated examples with layout blocks and structural relations, such as heading hierarchy, truncated text, and tables.}\label{fig:data_pipeline}
\end{figure}

\subsection{Iterative Document Annotation}
To build high-quality document-level annotations, we adopt an iterative annotation pipeline, as illustrated in Fig.~\ref{fig:data_pipeline}. The overall process starts from model-based pre-annotation and is then progressively improved annotation completeness and correctness through human correction and multi-dimensional verification feedback. The following sections describe the annotation details.

\subsubsection{Document Structure Annotation}
\textbf{Structure pre-annotation.} We first apply PP-DocLayoutV2~\cite{paddleocrvl} to documents for layout detection, obtaining the bounding boxes, categories, and reading order of layout blocks. However, these pre-annotations are still incomplete. In particular, they do not provide fine-grained figure subcategories (e.g., seals, flowcharts, and charts), and they also lack annotations for truncated relations and document hierarchical structures, which are essential for benchmark evaluation. Therefore, further manual annotation is needed to correct the pre-annotation errors and add the missing labels, including figure subcategories and document logical structures.

\textbf{Human correction.} To refine the pre-annotations, we define three types of manual annotation for correction and quality verification as shown in Fig. \ref{fig:data_pipeline}: (1) \textit{\textbf{Layout block annotation}}, which records the page index, bounding box, and category of each block; the fine-grained layout categories are shown in Tab.~\ref{fig:supp_data_statistics}. (2) \textit{\textbf{Reading order annotation}}, which assigns a unique reading order to each block across the document, excluding headers, footers, aside texts, and page numbers. (3) \textit{\textbf{Structural relation annotation}}, which specifies the parent node of each block and marks truncation relations for truncated paragraphs and tables. These parent relations further define the hierarchical structure of the document; in Fig.~\ref{fig:data_pipeline}, the arrows on the right illustrate the heading hierarchy tree among layout blocks. Although both support document-level evaluation, our annotations are richer, while READoc only provides ground-truth Markdown without visual information or explicit truncation relations, making it less suitable for realistic document parsing scenarios.

\textbf{Verification.} After human correction, we verify the annotation results through both automatic code-based validation and manual inspection. For automatic validation, we develop Python code to detect structural inconsistencies in the annotations, including cyclic hierarchy relations, nonexistent parent nodes, and invalid truncation annotations. For manual verification, each corrected document is visualized back onto the original PDF and reviewed by three quality inspectors. A document is approved only when all three inspectors agree that the annotations are correct; otherwise, the inspection comments, together with the corresponding sample, are sent back to the annotators for further revision. This iterative correction-and-verification process improves annotation accuracy, reduces residual errors, and enhances consistency across the dataset. 

Finally, we present several statistics of the annotated results in Fig.~\ref{fig:data_statistics}. The distribution of maximum heading-tree depth shows that both shallow and deeply nested document hierarchies are well represented. Cross-page dependencies are also common in the dataset. Specifically, 27.2\% of truncated text groups span multiple pages. For truncated tables, most merged cases cover two pages, while a smaller portion extends to three or four pages. These statistics demonstrate the diversity and realism of MPDocBench-Parse for multi-page document parsing evaluation.

\subsubsection{Document Content Annotation}

\textbf{Content pre-annotation.} 
After obtaining the document structure annotations, we crop each non-figure layout block from the document for content annotation. Following prior benchmarks, these cropped blocks are divided into three groups: text, tables, and formulas. We then use three representative content recognition models, Qwen-VL~\cite{bailian}, PaddleOCR-VL-1.5~\cite{paddleocrvl15}, and Mineru2.5~\cite{mineru25}, to recognize the content of each block.
For text blocks, we design a filtering strategy to identify potentially unreliable recognition results. Specifically, we compute pairwise similarities among the predictions of the three models and average them to obtain a recognition consistency score, denoted as $Sim_{avg}$. We use 1 - \text{Normalized Edit Distance~\cite{ned} (NED)} as the similarity metric. If $Sim_{avg}\leq0.95$, the block is considered ambiguous and sent for manual annotation and verification; 
otherwise, we use an advanced MLLM~\cite{gemini3pro} to refine the original labels, following DianJin-OCR-R1~\cite{dianjinocr}, thereby obtaining improved labels (see App.~\ref{ocr_label_refinement} for details).
For table and formula blocks, we initialize the pseudo labels with PaddleOCR-VL-1.5 predictions, with tables in HTML format and formulas in LaTeX format, and then manually correct and verify them to obtain the final ground truth.

\textbf{Human correction and verification.} Compared with structure annotation, content correction is simpler. Since each sample corresponds to a cropped image of a single layout block, annotators can directly revise the initial prediction based on the block image, the recognized content, and the task type. For complex formulas and tables, the revised results can be rendered using \textit{latexlive}\footnote{https://www.latexlive.com/} and \textit{onlineviewer}\footnote{https://html.onlineviewer.net/}, respectively, and compared against the original images. 
For tables requiring merging, annotators jointly review candidate tables and initial predictions, then determine the final merged result based on semantics, header consistency, truncation, and merging type. The entire verification stage follows the same iterative correction-and-verification strategy described above.

\subsection{MPDocBench-Parse Evaluation}
Following OmniDocBench, we adopt NED for text recognition and reading order, CDM~\cite{cdm} for formula recognition, and Tree-Edit-Distance-based Similarity~\cite{teds} (TEDS) for table recognition. In addition to these tasks, we further evaluate four challenging aspects of realistic document parsing.

\textbf{Truncated Text Merging.} We identify all ground-truth truncated text groups based on the annotated layout type (\texttt{text\_block}) and structural relation (\texttt{truncated}), merge the corresponding text blocks into paragraph-level units, and match them with predicted text segments using bipartite matching. The similarity of each matched pair is measured by NED, and the final score is obtained by averaging over all matched pairs.

\textbf{Truncated Table Merging.} We construct each ground-truth merged table from the annotations (layout type: \texttt{table}; structural relation: \texttt{truncated}), match it with predicted HTML tables extracted from Markdown, and evaluate each matched pair using TEDS. The overall score is computed by averaging across all merged tables.

\textbf{Figure Extraction.} Ground-truth and predicted figure boxes are matched page by page using the IoU between bounding boxes. Let \(M_{\text{match}}\), \(M_{\text{pred}}\), and \(M_{\text{gt}}\) denote the numbers of matched, predicted, and ground-truth figures in a document. The figure extraction F1 score is defined as $
\mathrm{Figure}_{\mathrm{F1}}=\frac{2M_{\text{match}}}{M_{\text{pred}}+M_{\text{gt}}}.$ 
The final performance is obtained by averaging the F1 score over all documents.

\textbf{Heading Hierarchy Recovery.} We convert the predicted and ground-truth headings into tree structures. The predicted document tree is derived from the number of Markdown heading prefix symbols ``\#'', while the ground-truth tree is built from the annotated relation (\texttt{parent\_son}) between headings. The similarity between the two trees is measured by TEDS, and the final score is averaged over all documents. {The corresponding tree construction process is provided in the App.~\ref{evaluation_detials}.}

All metrics are computed objectively by matching model predictions with annotated ground truth and applying the corresponding task-specific scoring functions. To obtain the overall result, we aggregate the scores from all eight tasks. For tasks evaluated by NED, we use $(1-\mathrm{NED})\times100$ since lower NED indicates better performance. The final overall score is the average over the eight task scores.

\section{Experiments}
In this section, we systematically evaluate current document parsing models on MPDocBench-Parse. We first describe the experimental setup and then present the overall evaluation results. We further provide fine-grained analyses of individual tasks. These experiments reveal the behavior of current models and the key bottlenecks in realistic multi-page parsing, thereby offering valuable insights for future document parsing research.

\subsection{Experimental Setups}
To comprehensively evaluate current document parsing models in multi-page scenarios, we benchmark specialized document parsing models and general-purpose vision-language models on MPDocBench-Parse. The specialized parsers are divided into two groups: pipeline-based models~\cite{glmocr,paddleocrvl15,mineru25,youtuparsing,monkeyocr,dolphinv2}, which perform layout analysis, content recognition, and structure recovery in stages, and end-to-end models~\cite{dotsmocr,fireredocr,dotsocr,deepseekocr2,ocrverse,logicsparsingv2,qianfanocr,chandraocr2}, which directly generate Markdown from document images. We also include several general VLMs~\cite{gemini31,bailian,chatgpt,qwen3vl,internvl35} to assess their document parsing capability under the same setting. For fair comparison, we adopt a unified input format and inference setting whenever possible, set the temperature to 0 to reduce output randomness, and normalize all outputs into a unified Markdown format for evaluation.

\begin{table*}[t]
  \centering
  \caption{Performance of document parsing methods on MPDocBench-Parse over eight evaluation metrics. T-Text and T-Table denote Truncated Text and Table Merging, respectively. ``-'' indicates that the model is unable to perform the task, resulting in a score of 0.
}
  \vspace{-7pt}
  \renewcommand{\arraystretch}{0.85}
  \resizebox{1.0\textwidth}{!}{
    \begin{tabular}{c|c|c c c c c c c c}
    \toprule
    \textbf{Model} & \textbf{Overall}$\uparrow$ & \textbf{Text\textsuperscript{Edit}}$\downarrow$ &
    \textbf{T-Text\textsuperscript{Edit}}$\downarrow$ &\textbf{Formula\textsuperscript{CDM}}$\uparrow$ & \textbf{Table\textsuperscript{TEDS}}$\uparrow$ &  
    \textbf{T-Table\textsuperscript{TEDS}}$\uparrow$ & 
    \textbf{Figure\textsuperscript{F1}}$\uparrow$ & 
    \textbf{Read Order\textsuperscript{Edit}}$\downarrow$ & 
    \textbf{Heading\textsuperscript{TEDS}}$\uparrow$ \\ 
    \midrule
\multicolumn{10}{l}{\textbf{\textit{Pipeline-based Specialized VLMs}}} \\
\midrule
GLM-OCR~\cite{glmocr} & 74.70 & 0.060 & 0.314 & 85.25 & 82.41 & 63.10 & 71.90 & 0.125 & 44.79 \\

PaddleOCR-VL-1.5~\cite{paddleocrvl15} & \textbf{80.80} & 0.047 & \textbf{0.147} & 87.86 & 83.54 & 83.07 & \textbf{75.19} & 0.106 & 46.78  \\

MinerU2.5~\cite{mineru25} & 76.70 & 0.059 & 0.328 & 81.01 & \textbf{85.87} & \textbf{88.14} & 72.44 & 0.120 & 36.90 \\

Youtu-Parsing~\cite{youtuparsing} & 74.16 & 0.093 & 0.353 & 87.19 & 84.82 & 63.88 & 71.56 & 0.132 & 43.64 \\

MonkeyOCR-pro-3B~\cite{monkeyocr} & 74.37 & 0.056 & 0.300 & 88.59 & 78.57 & 61.17 & 73.79 & 0.120 & 40.34 \\

Dolphin-v2~\cite{dolphinv2} & 73.14 & 0.104 & 0.334 & 80.38 & 83.84 & 64.28 & 64.15 & 0.137 & 50.04 \\

\midrule
\multicolumn{10}{l}{\textbf{\textit{End-to-End Specialized VLMs}}} \\
\midrule

dots.mocr~\cite{dotsmocr} & 72.96 & 0.071 & 0.298 & 86.32 & 81.60 & 62.16 & 67.90 & 0.111 & 33.68 \\

FireRed-OCR~\cite{fireredocr} & 69.32 & \textbf{0.041} & 0.178 & \textbf{89.96} & 81.65 & 62.62 & - & 0.086 & \textbf{50.84} \\

dots.ocr~\cite{dotsocr} & 74.17 & 0.079 & 0.303 & 86.15 & 83.18 & 60.68 & 67.82 & 0.114 & 45.12 \\

DeepSeek-OCR2~\cite{deepseekocr2} & 75.64 & 0.070 & 0.284 & 84.31 & 81.07 & 63.01 & 73.25 & 0.106 & 49.53 \\

OCRVerse~\cite{ocrverse} & 64.49 & 0.103 & 0.291 & 87.14 & 83.83 & 63.96 & - & 0.150 & 35.45 \\

Logics-Parsing-v2~\cite{logicsparsing} & 74.57 & 0.048 & 0.312 & 87.12 & 83.76 & 63.86 & 71.69 & \textbf{0.085} & 34.62 \\

Qianfan-OCR~\cite{qianfanocr} & 71.57 & 0.100 & 0.474 & 88.74 & 82.87 & 62.41 & 58.09 & 0.113 & 49.19 \\

ChandraOCR 2~\cite{chandraocr2} & 74.72 & 0.099 & 0.288 & 87.01 & 84.51 & 64.67 & 64.94 & 0.133 & 48.64 \\
\midrule
\multicolumn{10}{l}{\textbf{\textit{General VLMs}}} \\
\midrule
Gemini-3.1-pro-preview~\cite{gemini31} & 71.79 & 0.070 & 0.232 & 88.37 & 81.68 & 61.26 & 59.08 & 0.127 & 26.94 \\

ChatGPT-5.2-2025-12-11~\cite{chatgpt} & 65.66 & 0.111 & 0.372 & 84.33 & 78.98 & 58.71 & 30.97 & 0.169 & 37.50 \\

Qwen3.6-plus~\cite{bailian} & 72.14 & 0.096 & 0.245 & 88.77 & 82.99 & 60.50 & 64.80 & 0.179 & 32.06 \\

Qwen3-VL-235B~\cite{qwen3vl} & 73.98 & 0.087 & 0.187 & 84.71 & 81.35 & 61.25 & 63.25 & 0.136 & 42.34 \\

InternVL-3.5-38B~\cite{internvl35} & 57.01 & 0.133 & 0.515 & 84.30 & 69.86 & 51.69 & 7.97 & 0.199 & 27.01 \\
\bottomrule
\end{tabular}%
}
\label{tab:main_results}
\vspace{-10pt}
\end{table*}

\subsection{Main Evaluation Results}
As shown in Tab.~\ref{tab:main_results}, the best specialized VLM outperforms the best general VLM by a noticeable margin of 6.82\%, achieving a score of 80.80\% compared with 73.98\%. 
Beyond the overall ranking, the benchmark results further provide several important insights into the strengths and limitations of current models, which we summarize as follows:

\textbf{1. Textual content recognition is no longer the primary bottleneck.}
As shown in Tab.~\ref{tab:main_results}, current document parsing models achieve strong performance on fundamental tasks such as text, table, and formula recognition. In particular, text recognition results are consistently strong across models, with relatively small performance gaps; the top-performing models, \ie FireRed-OCR, Logics-Parsing-V2, and PaddleOCR-VL-1.5, all achieve accuracy above 95\%.

\textbf{2. Content semantic continuity remains a major challenge.}
Although existing models can accurately recognize textual content in documents, they often fail to effectively merge semantically continuous text or table segments that are split by complex layouts or page breaks in real-world scenarios. 
This lack of parsing capability leads to severe performance degradation on truncated text and table merging tasks. 
Nevertheless, PaddleOCR-VL-1.5 and MinerU2.5 explicitly take this issue into account and adopt corresponding merging strategies, achieving promising results.

\textbf{3. Figure extraction still has room for improvement.}
Figure extraction lags far behind textual content recognition according to the results in Tab.~\ref{tab:main_results}, revealing persistent weaknesses in visual region perception and precise localization. 
This issue is particularly evident in end-to-end parsing VLMs~\cite{qianfanocr,dotsocr,dotsmocr,bailian,dolphinv2,internvl35}, which are often unstable in recovering figure boundaries and prone to missed detections, shifted boxes, or confusion with other visual elements. 
We believe this issue likely arises from two factors: first, these parsing VLMs may naturally inherit the well-known weaknesses of MLLMs~\cite{zhang2025mllms,liu2025spatial} on visual grounding and localization tasks; second, their training often lacks effective supervision for visual content, further limiting precise figure extraction.

\textbf{4. Heading hierarchy recovery remains a broadly challenging problem.}
Nearly all models perform poorly on heading hierarchy recovery, with especially large performance gaps across document types. 
The results in Tab.~\ref{tab:heading_domain} show that for documents with regular structures and explicit heading levels, \ie academic papers, government documents, and exam papers, models are generally more capable of recovering the document heading tree. 
In contrast, documents such as magazines, books, newspapers, prospectuses, and reports often involve complex layouts, inconsistent heading styles, or hierarchical cues that depend more on semantic understanding than on visual patterns, making the task substantially more difficult.
{Figures~\ref{fig:supp_benchsample1}-\ref{fig:supp_benchsample3} present several examples from different document types.}

\textbf{5. Specialized VLMs outperform General VLMs.} 
As shown in Tab.~\ref{tab:main_results}, specialized document parsing models generally outperform general VLMs overall, and no general VLM ranks first on any of the eight tasks. This suggests that specialized models better fit document parsing through domain-specific data and better meet practical needs. Although general VLMs offer stronger multimodal understanding and instruction-following abilities, they are not always the most suitable choice for this highly structured task, especially in terms of training efficiency and deployment cost.

\begin{table*}[t]
  \centering
  \caption{Comparison of document parsing methods for heading hierarchy recovery across different document types in MPDocBench-Parse.
}
\vspace{-7pt}
  \renewcommand{\arraystretch}{0.85}
  \resizebox{1.0\textwidth}{!}{
    \begin{tabular}{c|c| c c c c c c c c c c c c c c c}
    \toprule
    \makecell{\textbf{Model}} & 
    \makecell{\textbf{Overall}} & 
    \makecell{\textbf{Annual}\\\textbf{Report}} & 
    \makecell{\textbf{Book}} &
    \makecell{\textbf{Financial}\\\textbf{Report}} &
    \makecell{\textbf{Exam}\\\textbf{Paper}} &
    \makecell{\textbf{Government}} & 
    \makecell{\textbf{National}\\\textbf{Standard}} & 
    \makecell{\textbf{Handwritten}} &
    \makecell{\textbf{Magazine}} & 
    \makecell{\textbf{Guidance}} &
    \makecell{\textbf{Newspaper}} &   
    \makecell{\textbf{Academic}\\\textbf{Paper}} &
    \makecell{\textbf{PPT\&}\\\textbf{Slides}} &
    \makecell{\textbf{Textbook}} &
    \makecell{\textbf{Research}\\\textbf{Report}} & 
    \makecell{\textbf{Prospectus}} \\
    \midrule
\multicolumn{16}{l}{\textbf{\textit{Pipeline-based Specialized VLMs}}} \\
\midrule
GLM-OCR~\cite{glmocr} & 44.79 & 46.25 & 43.26 & 46.49 & 67.26 & 57.09 & 27.93 & 50.21 & 22.68 & 40.62 & 42.01 & 68.76 & 31.28 & 26.12 & 46.19 & 36.07 \\

PaddleOCR-VL-1.5~\cite{paddleocrvl15} & 46.78 & 43.05 & 44.48 & \textbf{64.26} & 65.20 & 51.16 & 41.22 & 47.53 & 26.17 & 45.08 & 37.06 & \textbf{77.50} & 36.47 & 31.68 & 42.30 & 41.42 \\

MinerU2.5~\cite{mineru25} & 36.90 & 32.82 & 31.73 & 26.53 & 60.75 & 63.81 & 23.94 & 50.53 & 20.20 & 36.51 & 24.28 & 57.08 & 29.81 & 18.87 & 37.79 & 20.40 \\

Youtu-Parsing~\cite{youtuparsing} &  44.28 & 42.85 & 42.19 & 37.13 & 65.99 & 73.75 & 41.86 & 47.44 & \textbf{28.93} & 46.47 & 25.20 & 45.12 & 38.08 & 26.87 & 42.16 & 42.61\\

MonkeyOCR-pro-3B~\cite{monkeyocr} & 40.34 & 39.10 & 36.31 & 30.64 & 56.59 & 60.78 & 20.03 & 50.80 & 23.82 & 41.60 & 30.55 & 58.68 & 31.16 & 25.06 & 43.49 & 35.49 \\

Dolphin-v2~\cite{dolphinv2} & 50.04 & \textbf{52.99} & \textbf{48.91} & 39.01 & 62.25 & 68.91 & 41.28 & 48.79 & 28.54 & 43.70 & 42.41 & 76.41 & 45.59 & 31.93 & 52.05 & 45.84 \\

\midrule
\multicolumn{16}{l}{\textbf{\textit{End-to-End Specialized VLMs}}} \\
\midrule

dots.mocr~\cite{dotsmocr} & 33.68 & 40.48 & 35.32 & 33.02 & 49.41 & 45.43 & 24.85 & 27.41 & 16.85 & 36.56 & 5.97 & 55.06 & 10.82 & 25.06 & 37.49 & 31.05 \\

FireRed-OCR~\cite{fireredocr} & \textbf{50.84} & 44.06 & 46.70 & 41.27 & 71.96 & \textbf{74.45} & 37.47 & 48.14 & 28.25 & 49.55 & \textbf{44.02} & 75.94 & 32.39 & 35.51 & 56.41 & 48.64 \\

dots.ocr~\cite{dotsocr} & 45.12 & 50.63 & 39.69 & 45.86 & 56.62 & 66.94 & \textbf{52.74} & 40.70 & 28.06 & 42.78 & 26.83 & 57.96 & 34.22 & 29.07 & 52.15 & 41.00 \\

DeepSeek-OCR2~\cite{deepseekocr2} & 49.53 & 47.76 & 38.43 & 42.75 & \textbf{71.99} & 71.95 & 36.33 & 48.65 & 26.80 & 43.50 & 42.36 & 68.62 & 33.05 & 36.46 & 53.71 & \textbf{57.09} \\

OCRVerse~\cite{ocrverse} & 35.45 & 30.33 & 25.89 & 34.74 & 56.38 & 40.86 & 36.64 & 44.97 & 18.20 & 27.40 & 19.09 & 56.58 & 22.18 & 27.30 & 46.71 & 29.43 \\

Logics-Parsing-v2~\cite{logicsparsing} & 34.62 & 29.70 & 28.98 & 24.84 & 58.04 & 60.89 & 20.85 & 28.26 & 16.27 & 36.24 & 25.07 & 56.35 & 32.30 & 19.62 & 32.27 & 26.06 \\

Qianfan-OCR~\cite{qianfanocr} & 49.19 & 46.40 & 41.44 & 42.60 & 67.24 & 66.25 & 32.50 & 53.64 & 26.88 & 45.20 & 23.30 & 75.09 & 35.85 & \textbf{36.86} & \textbf{61.81} & 53.23 \\

ChandraOCR 2~\cite{chandraocr2} & 48.64 & 48.94 & 42.44 & 42.46 & 67.34 & 71.18 & 47.60 & \textbf{56.08} & 27.40 & \textbf{52.29} & 27.46 & 61.00 & 45.49 & 31.53 & 52.29 & 44.37 \\
\midrule
\multicolumn{16}{l}{\textbf{\textit{General VLMs}}} \\
\midrule
Gemini-3.1-pro-preview~\cite{gemini31} & 26.94 & 20.41 & 23.11 & 21.02 & 50.54 & 36.61 & 14.95 & 33.72 & 25.71 & 26.55 & 24.67 & 18.62 & 34.99 & 29.21 & 18.88 & 20.66 \\

ChatGPT-5.2-2025-12-11~\cite{chatgpt} & 37.50 & 41.34 & 38.89 & 46.49 & 40.57 & 50.03 & 34.48 & 44.34 & 20.79 & 37.00 & 23.52 & 56.55 & 34.95 & 24.54 & 36.82 & 28.77 \\

Qwen3.6-plus~\cite{bailian} & 32.06 & 31.58 & 31.17 & 19.73 & 55.15 & 49.40 & 22.71 & 31.08 & 20.41 & 24.23 & 17.69 & 29.31 & \textbf{45.86} & 31.07 & 25.90 & 31.33 \\

Qwen3-VL-235B~\cite{qwen3vl} & 42.34 & 49.71 & 37.14 & 50.85 & 59.23 & 58.60 & 42.23 & 43.36 & 22.08 & 42.05 & 32.22 & 60.03 & 42.15 & 25.29 & 35.53 & 30.57 \\

InternVL-3.5-38B~\cite{internvl35} & 27.01 & 24.25 & 26.42 & 31.95 & 54.91 & 49.81 & 11.78 & 34.68 & 14.36 & 25.25 & 11.88 & 31.93 & 32.04 & 13.41 & 19.33 & 15.76 \\
\bottomrule
\end{tabular}%
}
\label{tab:heading_domain}
\vspace{-10pt}
\end{table*}

\begin{table*}[t]
  \centering
  \caption{Detailed comparison of truncated text and table merging. W-P and C-P denote within-page and cross-page merging groups, respectively. F1 measures whether a model correctly merges two truncated blocks (text or table) that belong to the same merged group. The columns under Heading Hierarchy Depth report the scores on documents with different hierarchy tree depths.}
  \vspace{-7pt}
  \renewcommand{\arraystretch}{0.85}
  \resizebox{1.0\textwidth}{!}{
    \begin{tabular}{c|c c c c c|c c c c c c c| c c c c}
\toprule
\multirow{2}{*}{\textbf{Model}}
& \multicolumn{5}{c|}{\textbf{Truncated Text Merging}\rule{0pt}{2.0ex}}
& \multicolumn{7}{c|}{\textbf{Truncated Table Merging}\rule{0pt}{2.0ex}} &
\multicolumn{4}{c}{\textbf{Heading Hierarchy Depth}\rule{0pt}{2.0ex}} \\
\cline{2-17}
& \textbf{W-P}\rule{0pt}{2.4ex} & \textbf{C-P} & \textbf{Overall} & \textbf{W-P F1} & \textbf{C-P F1} &  \textbf{1-page}\textbf{}
& \textbf{2-page} & \textbf{3-page} & \textbf{4-page} & \textbf{Overall} & \textbf{W-P F1} & \textbf{C-P F1} & \textbf{$\le2$} & \textbf{$\le4$} & \textbf{$\le6$} & \textbf{$\ge7$} \\
\midrule
\multicolumn{8}{l}{\textbf{\textit{Pipeline-Based Specialized VLMs}}} \\
\midrule

GLM-OCR~\cite{glmocr} & 0.323 & 0.288 & 0.314 & 10.14 & - & 57.21 & 66.73 & 52.94 & 31.94 & 63.10 & - & - & 88.23 & 87.32 & 85.71 & 82.59 \\

PaddleOCR-VL-1.5~\cite{paddleocrvl15} & \textbf{0.095} & 0.286 & \textbf{0.147} & \textbf{72.88} & - & 54.12 & 83.88 & \textbf{93.00} & \textbf{96.92} & 83.07 & - & 86.73 & \textbf{91.38} & \textbf{88.23} & \textbf{86.76} & \textbf{84.46}  \\
    
MinerU2.5~\cite{mineru25}  & 0.332 & 0.315 & 0.328 & - & - & 54.86 & \textbf{91.95} & 84.76 & 96.35 & \textbf{88.14} & - & \textbf{87.88} & 89.20 & 85.52 & 84.10 & 82.26 \\

Youtu-Parsing~\cite{youtuparsing} & 0.371 & 0.305 & 0.353 & 7.70 & - & 65.44 & 67.16 & 51.41 & 31.84 & 63.88 & - & - & 90.71 & 86.55 & 84.82 & 81.69
 \\

MonkeyOCR-pro-3B~\cite{monkeyocr} & 0.306 & 0.286 & 0.300 & - & - & 49.35 & 65.04 & 53.01 & 31.38 & 61.17 & - & - & 87.94 & 86.67 & 85.05 & 82.37 \\

Dolphin-v2~\cite{dolphinv2} & 0.346 & 0.301 & 0.334 & - & - & 68.30 & 67.33 & 51.72 & 31.70 & 64.28 & - & - & 88.49 & 85.37 & 85.11 & 82.17 \\

\midrule
\multicolumn{8}{l}{\textbf{\textit{End-to-End Specialized VLMs}}} \\
\midrule

dots.mocr~\cite{dotsmocr} & 0.303 & 0.286 & 0.298 & 14.02 & - & 60.23 & 65.53 & 50.90 & 32.06 & 62.16 & - & - & 88.00 & 85.67 & 83.62 & 80.93 \\

FireRed-OCR~\cite{fireredocr} & 0.138 & \textbf{0.283} & 0.178 & 67.80 & - & 53.26 & 66.87 & 50.96 & 31.68 & 62.62 & - & - & 85.76 & 83.27 & 81.27 & 77.06 \\

dots.ocr~\cite{dotsocr}  & 0.309 & 0.285 & 0.303 & 14.09 & - & 60.22 & 63.63 & 49.78 & 35.18 & 60.68 & - & - & 88.28 & 86.94 & 85.42 & 82.37 \\

DeepSeek-OCR2~\cite{deepseekocr2} & 0.283 & 0.287 & 0.284 & 40.07 & - & 60.15 & 66.69 & 49.83 & 35.63 & 63.01 & - & - & 90.97 & 87.68 & 86.21 & 83.66 \\

OCRVerse~\cite{ocrverse} & 0.292 & 0.291 & 0.291 & 9.52 & - & 64.66 & 67.54 & 50.22 & 32.21 & 63.96 & - & - & 82.83 & 79.91 & 77.39 & 73.98 \\

Logics-Parsing-v2~\cite{logicsparsing} & 0.322 & 0.286 & 0.312 & 7.93 & - & 64.50 & 67.90 & 47.50 & 32.01 & 63.86 & - & - & 88.58 & 85.77 & 84.79 & 82.65 \\

Qianfan-OCR~\cite{qianfanocr} & 0.544 & 0.288 & 0.474 & 24.52 & - & 64.20 & 65.27 & 51.68 & 32.71 & 62.41 & - & - & 81.89 & 87.68 & 84.96 & 82.06  \\

ChandraOCR 2~\cite{chandraocr2} & 0.286 & 0.294 & 0.288 & 10.03 & - & \textbf{72.51} & 67.34 & 51.89 & 32.35 & 64.67 & - & - & 88.90 & 87.26 & 85.40 & 82.44 \\

\midrule
\multicolumn{8}{l}{\textbf{\textit{General VLMs}}} \\
\midrule

Gemini-3.1-pro-preview~\cite{gemini31} & 0.209 & 0.293 & 0.232 & 26.54 & - & 65.56 & 63.86 & 50.35 & 32.02 & 61.26 & - & - & 88.40 & 84.22 & 81.75 & 78.66 \\

ChatGPT-5.2-2025-12-11~\cite{chatgpt} & 0.394 & 0.312 & 0.372 & 24.36 & - & 67.73 & 60.82 & 46.99 & 32.73 & 58.71 & \textbf{33.33} & - & 77.08 & 82.57 & 80.55 & 76.62 \\

Qwen3.6-plus~\cite{bailian} & 0.182 & 0.414 & 0.245 & 27.87 & - & 55.90 & 64.48 & 46.82 & 32.53 & 60.50 & - & - & 86.10 & 84.16 & 82.18 & 78.12  \\

Qwen3-VL-235B~\cite{qwen3vl} & 0.147 & 0.295 & 0.187 & 40.15 & - & 56.89 & 64.65 & 50.94 & 32.53 & 61.25 & - & - & 86.72 & 85.24 & 84.81 & 81.67 \\

InternVL-3.5-38B~\cite{internvl35} & 0.588 & 0.321 & 0.515 & 10.96 & - & 47.33 & 54.23 & 44.75 & 30.06 & 51.69 & - & - & 76.20 & 78.43 & 75.06 & 71.07 \\
    \bottomrule
    \end{tabular}%
  }
  \label{tab:ttext_cptable_results}
  \vspace{-15pt}
\end{table*}

\subsection{Fine-grained Analysis}

\begin{figure}[t] 
  \centering
  \begin{minipage}[t]{0.55\textwidth}
    \centering
    \captionof{table}{Comparison of figure extraction performance across different figure subcategories.}
    \label{tab:figure_subtypes}
    \vspace{-7pt}
    \renewcommand{\arraystretch}{1.0}
    \resizebox{\linewidth}{!}{
      \begin{tabular}{c|c| c c c c c c c c c}
      \toprule
      \makecell{\textbf{Model}} & 
      \makecell{\textbf{Overall}} & 
      \makecell{\textbf{Chart}} & 
      \makecell{\textbf{Seal}} &
      \makecell{\textbf{Image} \textbf{Option}} &
      \makecell{\textbf{Rich-text} \textbf{Image}} &
      \makecell{\textbf{Flowchart}} & 
      \makecell{\textbf{Molecular} \textbf{Diagram}} &
      \makecell{\textbf{Product} \textbf{Image}} &
      \makecell{\textbf{UI} \textbf{Screenshot}} &
      \makecell{\textbf{Others}} \\
      \midrule
      \multicolumn{11}{l}{\textbf{\textit{Pipeline-based Specialized VLMs}}} \\
      \midrule
      GLM-OCR~\cite{glmocr} & 71.90 & 80.66 & 2.84 & 80.91 & 50.63 & 74.24 & 77.13 & 77.43 & 81.63 & 74.59 \\
      
      PaddleOCR-VL-1.5~\cite{paddleocrvl15} & \textbf{75.19} & 82.86 & \textbf{69.56} & 76.57 & 51.61 & 76.36 & 79.43 & 75.45 & 84.20 & 76.08 \\
      MinerU2.5~\cite{mineru25} & 72.44 & 82.87 & 44.43 & 81.35 & 48.87 & \textbf{80.83} & 79.73 & 80.30 & 84.55 & 74.91 \\
      Youtu-Parsing~\cite{youtuparsing} & 71.86 & 80.35 & 33.66 & 81.28 & 53.37 & 79.12 & 80.50 & 74.61 & 85.41 & 73.40 \\
      MonkeyOCR-pro-3B~\cite{monkeyocr} & 73.79 & 86.70 & 10.80 & 71.83 & \textbf{55.93} & 75.57 & 82.95 & \textbf{83.84} & \textbf{88.17} & \textbf{78.81} \\
      Dolphin-v2~\cite{dolphinv2}  & 64.15 & 70.49 & 33.90 & 61.24 & 44.54 & 70.64 & - & 62.67 & 77.30 & 66.72 \\
      \midrule
      \multicolumn{11}{l}{\textbf{\textit{End-to-End Specialized VLMs}}} \\
      \midrule
      dots.mocr~\cite{dotsmocr} & 67.90 & 70.53 & 69.05 & 73.71 & 44.47 & 69.70 & 73.07 & 71.34 & 77.09 & 68.63 \\
      dots.ocr~\cite{dotsocr} & 67.82 & 70.47 & 67.09 & 72.79 & 47.93 & 71.76 & 72.88 & 74.79 & 76.89 & 69.34 \\
      DeepSeek-OCR2~\cite{deepseekocr2} & 73.25 & \textbf{86.58} & 52.18 & 83.45 & 48.90 & 75.30 & 69.50 & 72.00 & 87.81 & 74.20 \\
      Logics-Parsing-v2~\cite{logicsparsing} & 71.69 & 81.19 & 55.79 & 76.06 & 40.33 & 60.23 & 79.94 & 77.60 & 81.89 & 74.57 \\
      Qianfan-OCR~\cite{qianfanocr} & 58.09 & 7.62 & 20.24 & \textbf{85.14} & 34.75 & 78.01 & \textbf{83.37} & 78.37 & 85.50 & 72.16 \\
      ChandraOCR 2~\cite{chandraocr2} & 64.94 & 77.22 & 47.40 & 68.70 & 34.22 & 7.45 & - & 67.84 & 85.31 & 71.12 \\
      \midrule
      \multicolumn{11}{l}{\textbf{\textit{General VLMs}}} \\
      \midrule
      Gemini-3.1-pro-preview~\cite{gemini31} & 59.08 & 58.44 & 10.52 & 52.64 & 43.78 & 52.69 & - & 61.16 & 65.45 & 53.70 \\

      ChatGPT-5.2-2025-12-11~\cite{chatgpt} & 30.97 & 42.81 & 7.80 & - & 30.54 & 41.20 & - & 28.99 & 46.93 & 29.78  \\
      Qwen3.6-plus~\cite{bailian} & 64.80 & 79.03 & 20.17 & 21.92 & 40.90 & 69.37 & 76.26 & 78.58 & 84.29 & 64.81 \\
      Qwen3-VL-235B~\cite{qwen3vl} & 63.25 & 72.67 & 2.84 & 37.54 & 38.75 & 66.56 & - & 59.88 & 77.13 & 58.49  \\
      InternVL-3.5-38B~\cite{internvl35} & 7.97 & 9.42 & - & - & 3.75 & 7.91 & 16.30 & 7.61 & - & 7.48 \\
      \bottomrule
      \end{tabular}%
    }
  \end{minipage}\hfill 
  %
  \begin{minipage}[t]{0.42\textwidth}
    \centering
    \captionof{table}{Parsing quality vs. page numbers.}
    \label{tab:score_with_page_nums}
    \vspace{-7pt}
  \renewcommand{\arraystretch}{0.2}
  \resizebox{0.8\textwidth}{!}{
    \begin{tabular}{c| c c c c c }
    \toprule
    \makecell{\textbf{Model}} & 
    \makecell{\textbf{$\le4$}} & 
    \makecell{\textbf{$\le8$}} &
    \makecell{\textbf{$\le12$}} &
    \makecell{\textbf{$\le16$}} &
    \makecell{\textbf{$\ge17$}} \\ 
    \midrule
\multicolumn{6}{l}{\textbf{\textit{Pipeline-based Specialized VLMs}}} \\
\midrule

GLM-OCR~\cite{glmocr} & 87.58 & 84.57 & 81.77 & 82.37 & 84.65 \\
PaddleOCR-VL-1.5~\cite{paddleocrvl15} & \textbf{89.11} & \textbf{86.05} & \textbf{83.40} & \textbf{84.40} & \textbf{86.53} \\
MinerU2.5~\cite{mineru25} & 86.55 & 83.42 & 81.48 & 82.51 & 83.94 \\
Youtu-Parsing~\cite{youtuparsing} & 87.72 & 83.86 & 80.88 & 82.02 & 81.42 \\
MonkeyOCR-pro-3B~\cite{monkeyocr} & 87.03 & 84.36 & 80.81 & 82.32 & 84.92 \\
Dolphin-v2~\cite{dolphinv2} & 86.24 & 84.70 & 80.75 & 81.77 & 82.23 \\

\midrule
\multicolumn{6}{l}{\textbf{\textit{End-to-End Specialized VLMs}}} \\
\midrule

dots.mocr~\cite{dotsmocr} & 86.33 & 83.02 & 79.44 & 81.47 & 81.47 \\
FireRed-OCR~\cite{fireredocr} & 83.97 & 79.77 & 76.15 & 77.67 & 77.18 \\
dots.ocr~\cite{dotsocr} & 87.31 & 84.67 & 81.52 & 82.25 & 81.85 \\
DeepSeek-OCR2~\cite{deepseekocr2} & 88.60 & 85.50 & 82.76 & 83.57 & 84.54 \\
OCRVerse~\cite{ocrverse} & 80.73 & 76.42 & 73.35 & 73.80 & 73.24 \\
Logics-Parsing-v2~\cite{logicsparsing} & 86.56 & 84.21 & 81.87 & 82.52 & 83.44 \\
Qianfan-OCR~\cite{qianfanocr} & 86.06 & 84.02 & 80.88 & 81.82 & 84.74 \\
ChandraOCR 2~\cite{chandraocr2} & 87.72 & 84.52 & 81.94 & 82.52 & 81.44 \\

\midrule
\multicolumn{6}{l}{\textbf{\textit{General VLMs}}} \\
\midrule

Gemini-3.1-pro-preview~\cite{gemini31} & 85.39 & 81.04 & 77.65 & 78.95 & 77.59 \\
ChatGPT-5.2-2025-12-11~\cite{chatgpt} & 81.03 & 79.20 & 75.40 & 77.52 & 77.10 \\
Qwen3.6-plus~\cite{bailian} & 84.70 & 81.26 & 76.97 & 77.74 & 77.38 \\
Qwen3-VL-235B~\cite{qwen3vl} & 85.65 & 83.92 & 80.68 & 82.49 & 80.54 \\
InternVL-3.5-38B~\cite{internvl35} & 77.81 & 73.51 & 70.24 & 71.72 & 71.92 \\

\bottomrule
\end{tabular}%
}
\end{minipage}
\vspace{-20pt}
\end{figure}

\textbf{Truncated text merging.}
We further analyze truncated text merging in within-page (W-P) and cross-page (C-P) merging settings in Tab.~\ref{tab:ttext_cptable_results}. 
Overall, all models perform poorly on cross-page paragraph merging, indicating that semantic continuity across pages remains difficult to capture. 
For within-page text merging, PaddleOCR-VL-1.5, FireRed-OCR, Qwen3.6-plus, and Qwen3-VL-235B perform relatively better than the other models. 
PaddleOCR-VL-1.5 benefits from heuristic merging strategies during text recognition, while the end-to-end parsing models may implicitly leverage semantic continuity to merge truncated text blocks.
We also report F1 score to evaluate whether two truncated text blocks that should be merged are correctly merged in the predicted Markdown output, separately for within-page and cross-page cases. 
The results show that nearly all models fail on cross-page truncated text merging, which helps explain the large edit distances on this task.

\textbf{Truncated table merging.}
Tab.~\ref{tab:ttext_cptable_results} also reports detailed results on truncated table merging. Except for ChatGPT-5.2, which achieves 33.33\% F1 on within-page merging, nearly all models fail on this task. For cross-page merging, only PaddleOCR-VL-1.5 and MinerU2.5 achieve strong results, as they explicitly merge tables from adjacent pages based on page-level parsing outputs and reading order. However, TEDS also has limitations for evaluating table merging: a model may obtain a high score as long as it identifies the correct table fragments to merge, even if the merged structure is incomplete or partially incorrect. Thus, the score may not fully reflect the true quality of table merging. Detailed case studies are provided in App.~\ref{evaluation_detials}.

\textbf{Document parsing performance across different heading hierarchy depths.} As shown in Tab.~\ref{tab:ttext_cptable_results}, most models achieve relatively strong performance on documents with shallow heading structures, while their overall scores gradually decline as the hierarchy depth increases. This trend suggests that accurate document parsing becomes more difficult when heading relations are more deeply nested. 
We attribute this phenomenon to the fact that deeper heading hierarchies usually correspond to more complex document organization, requiring stronger structural understanding and long-range dependency modeling, which remains challenging for current models.

{\textbf{Figure extraction across subcategories.}}
Figure extraction performance varies across subcategories in Tab.~\ref{tab:figure_subtypes}. Most models achieve relatively consistent results on categories with salient visual patterns and clear boundaries, \ie chart, image option, flowchart, molecular diagram, and UI screenshot. A likely reason is that these categories usually contain more distinctive visual structures, regular layouts, or strong semantic cues, making them easier for models to localize accurately. Moreover, these figure types are also more frequently seen in existing training corpora, which may further improve model robustness on such categories.
In contrast, model performance varies much more on seals and rich-text images (visualizations can be seen in Fig.~\ref{fig:supp_figure_extraction}). 
Seal figures are often small and visually ambiguous, while rich-text images are usually embedded in complex text-image mixed layouts with less explicit boundaries.
As a result, performance on these categories is less consistent and generally more sensitive to model capability.

\textbf{Document parsing performance across different page counts.} As shown in Tab.~\ref{tab:score_with_page_nums}, as the number of input pages increases, most models exhibit a certain performance drop, followed by a slight recovery. However, their final performance remains worse than under the setting with no more than 4 pages. In addition, the average performance gap between the \(\leq 4\) and \(\geq 17\) settings is about 3\% for pipeline-based specialized VLMs, compared with 4.9\% for end-to-end specialized VLMs and 6\% for general VLMs. This suggests that pipeline-based specialized VLMs are more stable in realistic multi-page document parsing scenarios, which hasn't been systematically explored in previous work.

\section{Conclusion}
We introduce MPDocBench-Parse, a practical benchmark for multi-page document parsing. Unlike existing benchmarks that mainly focus on single-page, text-centric evaluation, it is designed around four key requirements of practical parsing: multi-page end-to-end evaluation, broad document diversity, content fidelity, and logical structural correctness. We further establish an eight-dimensional evaluation protocol grounded in realistic parsing needs, covering critical gaps in current evaluation such as semantic continuity, multimodal visual content extraction, and document hierarchy recovery. Through comprehensive experiments on specialized parsing models and general-purpose VLMs, we show that, while current systems have achieved encouraging progress on basic content recognition, substantial room for improvement remains, particularly in challenging aspects of practical parsing such as semantic continuity recovery and hierarchical structure reconstruction. These findings suggest that existing benchmarks may not yet fully capture the challenges of realistic multi-page parsing scenarios. We hope MPDocBench-Parse will serve as a valuable foundation for future research, and we will open-source all annotated data and evaluation code.

{
    \small
    \bibliographystyle{unsrt}
    \bibliography{main}

@String(AAAI = {AAAI})

@InProceedings{PubLayNet,
    author    = {Zhong Xu and Jianbin Tang and Antonio Jimeno Yepes.},
    title     = {Publaynet: largest dataset ever for document layout analysis},
    booktitle = {2019 International conference on document analysis and recognition},
    month     = {},
    year      = {2019},
    pages     = {1015-1022}
}

@InProceedings{PubTabNet,
    author    = {Zhong, Xu and ShafieiBavani, Elaheh and Jimeno Yepes, Antonio},
    title     = {Image-based table recognition: data, model, and evaluation},
    booktitle = {European conference on computer vision},
    month     = {},
    year      = {2020},
    pages     = {564--580}
}

@InProceedings{M6Doc,
    author    = {Cheng, Hiuyi and Zhang, Peirong and Wu, Sihang and Zhang, Jiaxin and Zhu, Qiyuan and Xie, Zecheng and Li, Jing and Ding, Kai and Jin, Lianwen},
    title     = {M6Doc: A Large-Scale Multi-Format, Multi-Type, Multi-Layout, Multi-Language, Multi-Annotation Category Dataset for Modern Document Layout Analysis},
    booktitle = {Proceedings of the IEEE/CVF Conference on Computer Vision and Pattern Recognition},
    month     = {June},
    year      = {2023},
    pages     = {15138-15147}
}

@InProceedings{RobustReading,
    author    = {Karatzas, Dimosthenis and Gomez-Bigorda, Lluis and Nicolaou, Anguelos and Ghosh, Suman and Bagdanov, Andrew and Iwamura, Masakazu and Matas, Jiri and Neumann, Lukas and Chandrasekhar, Vijay Ramaseshan and Lu, Shijian and Shafait, Faisal and Uchida, Seiichi and Valveny, Ernest},
    title     = {ICDAR 2015 competition on Robust Reading},
    booktitle = {2015 13th International Conference on Document Analysis and Recognition},
    month     = {},
    year      = {2015},
    pages     = {1156-1160}
}

@InProceedings{DocLayNet,
    author    = {Pfitzmann, Birgit and Auer, Christoph and Dolfi, Michele and Nassar, Ahmed S and Staar, Peter},
    title     = {Doclaynet: A large human-annotated dataset for document-layout segmentation},
    booktitle = {Proceedings of the 28th ACM SIGKDD conference on knowledge discovery and data mining},
    month     = {},
    year      = {2022},
    pages     = {3743--3751}
}

@inproceedings{Tablebank,
  title={Tablebank: Table benchmark for image-based table detection and recognition},
  author={Li, Minghao and Cui, Lei and Huang, Shaohan and Wei, Furu and Zhou, Ming and Li, Zhoujun},
  booktitle={Proceedings of the Twelfth Language Resources and Evaluation Conference},
  pages={1918--1925},
  year={2020}
}

@inproceedings{Im2Latex-100K,
  title={Image-to-markup generation with coarse-to-fine attention},
  author={Deng, Yuntian and Kanervisto, Anssi and Ling, Jeffrey and Rush, Alexander M},
  booktitle={International Conference on Machine Learning},
  pages={980--989},
  year={2017},
  organization={PMLR}
}

@misc{wang2024unimernetuniversalnetworkrealworld,
      title={UniMERNet: A Universal Network for Real-World Mathematical Expression Recognition}, 
      author={Bin Wang and Zhuangcheng Gu and Guang Liang and Chao Xu and Bo Zhang and Botian Shi and Conghui He},
      year={2024},
      eprint={2404.15254},
      archivePrefix={arXiv},
      primaryClass={cs.CV},
      url={https://arxiv.org/abs/2404.15254}, 
}

@article{fox,
  title={Focus Anywhere for Fine-grained Multi-page Document Understanding},
  author={Liu, Chenglong and Wei, Haoran and Chen, Jinyue and Kong, Lingyu and Ge, Zheng and Zhu, Zining and Zhao, Liang and Sun, Jianjian and Han, Chunrui and Zhang, Xiangyu},
  journal={arXiv:2405.14295},
  year={2024}
}

@article{got2,
  title={General ocr theory: Towards ocr-2.0 via a unified end-to-end model},
  author={Wei, Haoran and Liu, Chenglong and Chen, Jinyue and Wang, Jia and Kong, Lingyu and Xu, Yanming and Ge, Zheng and Zhao, Liang and Sun, Jianjian and Peng, Yuang and others},
  journal={arXiv:2409.01704},
  year={2024}
}

@article{Nougat,
author = {Lukas Blecher and Guillem Cucurull and Thomas Scialom and Robert Stojnic},
title = {Nougat: Neural Optical Understanding for Academic Documents},
journal = {arXiv:2308.13418},
pages = {},
year = 2024
}

@inproceedings{readoc,
  title={Readoc: A unified benchmark for realistic document structured extraction},
  author={Li, Zichao and Abulaiti, Aizier and Lu, Yaojie and Chen, Xuanang and Zheng, Jia and Lin, Hongyu and Han, Xianpei and Jiang, Shanshan and Dong, Bin and Sun, Le},
  booktitle={Findings of the Association for Computational Linguistics: ACL 2025},
  pages={21889--21905},
  year={2025}
}

@article{olmocr,
  title={olmocr: Unlocking trillions of tokens in pdfs with vision language models},
  author={Poznanski, Jake and Rangapur, Aman and Borchardt, Jon and Dunkelberger, Jason and Huff, Regan and Lin, Daniel and Wilhelm, Christopher and Lo, Kyle and Soldaini, Luca},
  journal={arXiv preprint arXiv:2502.18443},
  year={2025}
}

@inproceedings{omnidocbench,
  title={Omnidocbench: Benchmarking diverse pdf document parsing with comprehensive annotations},
  author={Ouyang, Linke and Qu, Yuan and Zhou, Hongbin and Zhu, Jiawei and Zhang, Rui and Lin, Qunshu and Wang, Bin and Zhao, Zhiyuan and Jiang, Man and Zhao, Xiaomeng and others},
  booktitle={Proceedings of the IEEE/CVF Conference on Computer Vision and Pattern Recognition},
  pages={24838--24848},
  year={2025}
}

@inproceedings{dochienet,
  title={Dochienet: A large and diverse dataset for document hierarchy parsing},
  author={Xing, Hangdi and Cheng, Changxu and Gao, Feiyu and Shao, Zirui and Yu, Zhi and Bu, Jiajun and Zheng, Qi and Yao, Cong},
  booktitle={Proceedings of the 2024 Conference on Empirical Methods in Natural Language Processing},
  pages={1129--1142},
  year={2024}
}

@inproceedings{hrdoc,
  title={Hrdoc: Dataset and baseline method toward hierarchical reconstruction of document structures},
  author={Ma, Jiefeng and Du, Jun and Hu, Pengfei and Zhang, Zhenrong and Zhang, Jianshu and Zhu, Huihui and Liu, Cong},
  booktitle={Proceedings of the AAAI Conference on Artificial Intelligence},
  volume={37},
  number={2},
  pages={1870--1877},
  year={2023}
}

@inproceedings{layoutreader,
  title={Layoutreader: Pre-training of text and layout for reading order detection},
  author={Wang, Zilong and Xu, Yiheng and Cui, Lei and Shang, Jingbo and Wei, Furu},
  booktitle={Proceedings of the 2021 conference on empirical methods in natural language processing},
  pages={4735--4744},
  year={2021}
}

@article{scitsr,
  title={Complicated table structure recognition},
  author={Chi, Zewen and Huang, Heyan and Xu, Heng-Da and Yu, Houjin and Yin, Wanxuan and Mao, Xian-Ling},
  journal={arXiv preprint arXiv:1908.04729},
  year={2019}
}

@inproceedings{fintabnet,
  title={Global table extractor (gte): A framework for joint table identification and cell structure recognition using visual context},
  author={Zheng, Xinyi and Burdick, Douglas and Popa, Lucian and Zhong, Xu and Wang, Nancy Xin Ru},
  booktitle={Proceedings of the IEEE/CVF winter conference on applications of computer vision},
  pages={697--706},
  year={2021}
}

@inproceedings{multidocfusion,
  title={MultiDocFusion: Hierarchical and Multimodal Chunking Pipeline for Enhanced RAG on Long Industrial Documents},
  author={Shin, Joongmin and Park, Chanjun and Park, Jeongbae and Seo, Jaehyung and Lim, Heui-Seok},
  booktitle={Proceedings of the 2025 Conference on Empirical Methods in Natural Language Processing},
  pages={20996--21015},
  year={2025}
}

@article{bookrag,
  title={BookRAG: A Hierarchical Structure-aware Index-based Approach for Retrieval-Augmented Generation on Complex Documents},
  author={Wang, Shu and Zhou, Yingli and Fang, Yixiang},
  journal={arXiv preprint arXiv:2512.03413},
  year={2025}
}

@article{deepread,
  title={DeepRead: Document Structure-Aware Reasoning to Enhance Agentic Search},
  author={Li, Zhanli and Tian, Huiwen and Luo, Lvzhou and Cao, Yixuan and Luo, Ping},
  journal={arXiv preprint arXiv:2602.05014},
  year={2026}
}

@article{logicsparsing,
  title={Logics-parsing technical report},
  author={Chen, Xiangyang and Li, Shuzhao and Zhu, Xiuwen and Chen, Yongfan and Yang, Fan and Fang, Cheng and Qu, Lin and Xu, Xiaoxiao and Wei, Hu and Wu, Minggang},
  journal={arXiv preprint arXiv:2509.19760},
  year={2025}
}

@article{logicsparsingv2,
  title={Logics-Parsing-Omni Technical Report},
  author={An, Xin and Cai, Jingyi and Chen, Xiangyang and Liu, Huayao and Liu, Peiting and Wang, Peng and Yang, Bei and Zhu, Xiuwen and Chen, Yongfan and Hou, Baoyu and others},
  journal={arXiv preprint arXiv:2603.09677},
  year={2026}
}

@article{mmlongbenchdoc,
  title={Mmlongbench-doc: Benchmarking long-context document understanding with visualizations},
  author={Ma, Yubo and Zang, Yuhang and Chen, Liangyu and Chen, Meiqi and Jiao, Yizhu and Li, Xinze and Lu, Xinyuan and Liu, Ziyu and Ma, Yan and Dong, Xiaoyi and others},
  journal={Advances in Neural Information Processing Systems},
  volume={37},
  pages={95963--96010},
  year={2024}
}

@inproceedings{longdocurl,
  title={Longdocurl: a comprehensive multimodal long document benchmark integrating understanding, reasoning, and locating},
  author={Deng, Chao and Yuan, Jiale and Bu, Pi and Wang, Peijie and Li, Zhong-Zhi and Xu, Jian and Li, Xiao-Hui and Gao, Yuan and Song, Jun and Zheng, Bo and others},
  booktitle={Proceedings of the 63rd Annual Meeting of the Association for Computational Linguistics (Volume 1: Long Papers)},
  pages={1135--1159},
  year={2025}
}

@article{unidoc,
  title={Unidoc-bench: A unified benchmark for document-centric multimodal rag},
  author={Peng, Xiangyu and Qin, Can and Chen, Zeyuan and Xu, Ran and Xiong, Caiming and Wu, Chien-Sheng},
  journal={arXiv preprint arXiv:2510.03663},
  year={2025}
}

@inproceedings{docbench,
  title={Docbench: A benchmark for evaluating llm-based document reading systems},
  author={Zou, Anni and Yu, Wenhao and Zhang, Hongming and Ma, Kaixin and Cai, Deng and Zhang, Zhuosheng and Zhao, Hai and Yu, Dong},
  booktitle={Proceedings of the 4th International Workshop on Knowledge-Augmented Methods for Natural Language Processing},
  pages={359--373},
  year={2025}
}

@article{paddleocrv3,
  title={Paddleocr 3.0 technical report},
  author={Cui, Cheng and Sun, Ting and Lin, Manhui and Gao, Tingquan and Zhang, Yubo and Liu, Jiaxuan and Wang, Xueqing and Zhang, Zelun and Zhou, Changda and Liu, Hongen and others},
  journal={arXiv preprint arXiv:2507.05595},
  year={2025}
}

@inproceedings{docling,
  title={Docling: An Efficient Open-Source Toolkit for AI-driven Document Conversion},
  author={Livathinos, Nikos and Auer, Christoph and Lysak, Maxim and Nassar, Ahmed and Dolfi, Michele and Vagenas, Panos and Ramis, Cesar Berrospi and Omenetti, Matteo and Dinkla, Kasper and Kim, Yusik and others},
  booktitle={AAAI Conference on Artificial Intelligence},
  year={2025}
}

@inproceedings{layoutparser,
  title={Layoutparser: A unified toolkit for deep learning based document image analysis},
  author={Shen, Zejiang and Zhang, Ruochen and Dell, Melissa and Lee, Benjamin Charles Germain and Carlson, Jacob and Li, Weining},
  booktitle={International Conference on Document Analysis and Recognition},
  pages={131--146},
  year={2021},
  organization={Springer}
}

@inproceedings{da2023vision,
  title={Vision grid transformer for document layout analysis},
  author={Da, Cheng and Luo, Chuwei and Zheng, Qi and Yao, Cong},
  booktitle={Proceedings of the IEEE/CVF international conference on computer vision},
  pages={19462--19472},
  year={2023}
}

@article{monkeyocr,
  title={Monkeyocr: Document parsing with a structure-recognition-relation triplet paradigm},
  author={Li, Zhang and Liu, Yuliang and Liu, Qiang and Ma, Zhiyin and Zhang, Ziyang and Zhang, Shuo and Yang, Biao and Guo, Zidun and Zhang, Jiarui and Wang, Xinyu and others},
  journal={arXiv preprint arXiv:2506.05218},
  year={2025}
}

@article{mdpbench,
  title={MDPBench: A Benchmark for Multilingual Document Parsing in Real-World Scenarios},
  author={Li, Zhang and Lin, Zhibo and Liu, Qiang and Zhang, Ziyang and Zhang, Shuo and Guo, Zidun and Song, Jiajun and Zhang, Jiarui and Bai, Xiang and Liu, Yuliang},
  journal={arXiv preprint arXiv:2603.28130},
  year={2026}
}

@article{paddleocrvl,
  title={Paddleocr-vl: Boosting multilingual document parsing via a 0.9 b ultra-compact vision-language model},
  author={Cui, Cheng and Sun, Ting and Liang, Suyin and Gao, Tingquan and Zhang, Zelun and Liu, Jiaxuan and Wang, Xueqing and Zhou, Changda and Liu, Hongen and Lin, Manhui and others},
  journal={arXiv preprint arXiv:2510.14528},
  year={2025}
}

@article{paddleocrvl15,
  title={PaddleOCR-VL-1.5: Towards a Multi-Task 0.9 B VLM for Robust In-the-Wild Document Parsing},
  author={Cui, Cheng and Sun, Ting and Liang, Suyin and Gao, Tingquan and Zhang, Zelun and Liu, Jiaxuan and Wang, Xueqing and Zhou, Changda and Liu, Hongen and Lin, Manhui and others},
  journal={arXiv preprint arXiv:2601.21957},
  year={2026}
}

@article{mineru,
  title={Mineru: An open-source solution for precise document content extraction},
  author={Wang, Bin and Xu, Chao and Zhao, Xiaomeng and Ouyang, Linke and Wu, Fan and Zhao, Zhiyuan and Xu, Rui and Liu, Kaiwen and Qu, Yuan and Shang, Fukai and others},
  journal={arXiv preprint arXiv:2409.18839},
  year={2024}
}

@article{mineru25,
  title={Mineru2. 5: A decoupled vision-language model for efficient high-resolution document parsing},
  author={Niu, Junbo and Liu, Zheng and Gu, Zhuangcheng and Wang, Bin and Ouyang, Linke and Zhao, Zhiyuan and Chu, Tao and He, Tianyao and Wu, Fan and Zhang, Qintong and others},
  journal={arXiv preprint arXiv:2509.22186},
  year={2025}
}

@article{cdm,
  title={Cdm: A reliable metric for fair and accurate formula recognition evaluation},
  author={Wang, Bin and Wu, Fan and Ouyang, Linke and Gu, Zhuangcheng and Zhang, Rui and Xia, Renqiu and Zhang, Bo and He, Conghui},
  journal={arXiv preprint arXiv:2409.03643},
  volume={5},
  number={6},
  year={2024}
}

@inproceedings{ned,
  title={Binary codes capable of correcting deletions, insertions, and reversals},
  author={Levenshtein, Vladimir I and others},
  booktitle={Soviet physics doklady},
  volume={10},
  number={8},
  pages={707--710},
  year={1966},
  organization={Soviet Union}
}

@inproceedings{teds,
  title={Image-based table recognition: data, model, and evaluation},
  author={Zhong, Xu and ShafieiBavani, Elaheh and Jimeno Yepes, Antonio},
  booktitle={European conference on computer vision},
  pages={564--580},
  year={2020},
  organization={Springer}
}

@article{glmocr,
  title={Glm-ocr technical report},
  author={Duan, Shuaiqi and Xue, Yadong and Wang, Weihan and Su, Zhe and Liu, Huan and Yang, Sheng and Gan, Guobing and Wang, Guo and Wang, Zihan and Yan, Shengdong and others},
  journal={arXiv preprint arXiv:2603.10910},
  year={2026}
}

@article{fireredocr,
  title={FireRed-OCR Technical Report},
  author={Wu, Hao and Lou, Haoran and Li, Xinyue and Zhong, Zuodong and Sun, Zhaojun and Chen, Phellon and Zhou, Xuanhe and Zuo, Kai and Chen, Yibo and Tang, Xu and others},
  journal={arXiv preprint arXiv:2603.01840},
  year={2026}
}

@article{youtuparsing,
  title={Youtu-Parsing: Perception, Structuring and Recognition via High-Parallelism Decoding},
  author={Yin, Kun and Wu, Yunfei and Liu, Bing and Cai, Zhongpeng and Li, Xiaotian and Chen, Huang and Li, Xin and Cao, Haoyu and Liu, Yinsong and Jiang, Deqiang and others},
  journal={arXiv preprint arXiv:2601.20430},
  year={2026}
}

@article{dotsocr,
  title={dots. ocr: Multilingual document layout parsing in a single vision-language model},
  author={Li, Yumeng and Yang, Guang and Liu, Hao and Wang, Bowen and Zhang, Colin},
  journal={arXiv preprint arXiv:2512.02498},
  year={2025}
}

@article{qianfanocr,
  title={Qianfan-OCR: A Unified End-to-End Model for Document Intelligence},
  author={Dong, Daxiang and Zheng, Mingming and Xu, Dong and Luo, Chunhua and Zhuang, Bairong and Li, Yuxuan and He, Ruoyun and Wang, Haoran and Zhang, Wenyu and Wang, Wenbo and others},
  journal={arXiv preprint arXiv:2603.13398},
  year={2026}
}

@article{deepseekocr2,
  title={DeepSeek-OCR 2: Visual Causal Flow},
  author={Wei, Haoran and Sun, Yaofeng and Li, Yukun},
  journal={arXiv preprint arXiv:2601.20552},
  year={2026}
}

@article{dolphinv2,
  title={Dolphin-v2: Universal Document Parsing via Scalable Anchor Prompting},
  author={Feng, Hao and Shi, Wei and Zhang, Ke and Fei, Xiang and Liao, Lei and Yang, Dingkang and Du, Yongkun and Wu, Xuecheng and Tang, Jingqun and Liu, Yang and others},
  journal={arXiv preprint arXiv:2602.05384},
  year={2026}
}

@article{ocrverse,
  title={OCRVerse: Towards Holistic OCR in End-to-End Vision-Language Models},
  author={Zhong, Yufeng and Chen, Lei and Zhao, Xuanle and Han, Wenkang and Zheng, Liming and Huang, Jing and Jiang, Deyang and Cao, Yilin and Ma, Lin and Zeng, Zhixiong},
  journal={arXiv preprint arXiv:2601.21639},
  year={2026}
}

@article{qwen3vl,
  title={Qwen3-vl technical report},
  author={Bai, Shuai and Cai, Yuxuan and Chen, Ruizhe and Chen, Keqin and Chen, Xionghui and Cheng, Zesen and Deng, Lianghao and Ding, Wei and Gao, Chang and Ge, Chunjiang and others},
  journal={arXiv preprint arXiv:2511.21631},
  year={2025}
}

@article{internvl35,
  title={Internvl3. 5: Advancing open-source multimodal models in versatility, reasoning, and efficiency},
  author={Wang, Weiyun and Gao, Zhangwei and Gu, Lixin and Pu, Hengjun and Cui, Long and Wei, Xingguang and Liu, Zhaoyang and Jing, Linglin and Ye, Shenglong and Shao, Jie and others},
  journal={arXiv preprint arXiv:2508.18265},
  year={2025}
}

@article{dotsmocr,
  title={Multimodal OCR: Parse Anything from Documents},
  author={Zheng, Handong and Li, Yumeng and Zhang, Kaile and Xin, Liang and Zhao, Guangwei and Liu, Hao and Chen, Jiayu and Lou, Jie and Qiu, Jiyu and Fu, Qi and others},
  journal={arXiv preprint arXiv:2603.13032},
  year={2026}
}

@misc{gemini3pro,
  title={Gemini 3 Pro},
  author={{Google DeepMind}},
  year={2025},
  howpublished={\url{https://blog.google/innovation-and-ai/technology/developers-tools/gemini-3-pro-vision}}
}

@misc{gemini31,
  title={Gemini 3.1},
  author={{Google DeepMind}},
  year={2026},
  howpublished={\url{https://deepmind.google/models/gemini/pro/}}
}

@misc{chatgpt,
  title={ChatGPT},
  author={{OpenAI}},
  year={2025},
  howpublished={\url{https://chat.openai.com}},
}

@misc{claude,
  title={Claude},
  author={{Anthropic}},
  year={2025},
  howpublished={\url{https://www.anthropic.com/claude}},
}

@misc{bailian,
  title={TongYi},
  author={{TongYi}},
  year={2025},
  howpublished={\url{https://www.aliyun.com/product/tongyi}},
}

@misc{chandraocr2,
  author = {Chandra OCR 2},
  title = {Chandra OCR 2},
  year = {2025},
  howpublished = {\url{https://github.com/datalab-to/chandra}}
}

@article{docptbench,
  title={DocPTBench: Benchmarking End-to-End Photographed Document Parsing and Translation},
  author={Du, Yongkun and Chen, Pinxuan and Ying, Xuye and Chen, Zhineng},
  journal={arXiv preprint arXiv:2511.18434},
  year={2025}
}

@inproceedings{ccocr,
  title={Cc-ocr: A comprehensive and challenging ocr benchmark for evaluating large multimodal models in literacy},
  author={Yang, Zhibo and Tang, Jun and Li, Zhaohai and Wang, Pengfei and Wan, Jianqiang and Zhong, Humen and Liu, Xuejing and Yang, Mingkun and Wang, Peng and Bai, Shuai and others},
  booktitle={Proceedings of the IEEE/CVF International Conference on Computer Vision},
  pages={21744--21754},
  year={2025}
}

@article{yeh2026kohakurag,
  title={KohakuRAG: A simple RAG framework with hierarchical document indexing},
  author={Yeh, Shih-Ying and Ku, Yueh-Feng and Huang, Ko-Wei and Tu, Buu-Khang},
  journal={arXiv preprint arXiv:2603.07612},
  year={2026}
}

@article{zhang2026parsebench,
  title={ParseBench: A Document Parsing Benchmark for AI Agents},
  author={Zhang, Boyang and Acosta, Sebasti{\'a}n G and Carlson, Preston and Bron, Sacha and Doulcet, Pierre-Lo{\"\i}c and Suo, Simon},
  journal={arXiv preprint arXiv:2604.08538},
  year={2026}
}

@article{zhang2025mllms,
  title={Why do mllms struggle with spatial understanding? a systematic analysis from data to architecture},
  author={Zhang, Wanyue and Huang, Yibin and Xu, Yangbin and Huang, JingJing and Zhi, Helu and Ren, Shuo and Xu, Wang and Zhang, Jiajun},
  journal={arXiv preprint arXiv:2509.02359},
  year={2025}
}

@article{liu2025spatial,
  title={Spatial Reasoning in Multimodal Large Language Models: A Survey of Tasks, Benchmarks and Methods},
  author={Liu, Weichen and Xue, Qiyao and Wang, Haoming and Yin, Xiangyu and Yang, Boyuan and Gao, Wei},
  journal={arXiv preprint arXiv:2511.15722},
  year={2025}
}

@inproceedings{vllm,
  title={Efficient Memory Management for Large Language Model Serving with PagedAttention},
  author={Woosuk Kwon and Zhuohan Li and Siyuan Zhuang and Ying Sheng and Lianmin Zheng and Cody Hao Yu and Joseph E. Gonzalez and Hao Zhang and Ion Stoica},
  booktitle={Proceedings of the ACM SIGOPS 29th Symposium on Operating Systems Principles},
  year={2023}
}

@article{lmdeploy,
  title={Efficient Mixed-Precision Large Language Model Inference with TurboMind},
  author={Zhang, Li and Jiang, Youhe and He, Guoliang and Chen, Xin and Lv, Han and Yao, Qian and Fu, Fangcheng and Chen, Kai},
  journal={arXiv preprint arXiv:2508.15601},
  year={2025}
}

@inproceedings{transformers,
    title = "Transformers: State-of-the-Art Natural Language Processing",
    author = "Thomas Wolf and Lysandre Debut and Victor Sanh and Julien Chaumond and Clement Delangue and Anthony Moi and Pierric Cistac and Tim Rault and Rémi Louf and Morgan Funtowicz and Joe Davison and Sam Shleifer and Patrick von Platen and Clara Ma and Yacine Jernite and Julien Plu and Canwen Xu and Teven Le Scao and Sylvain Gugger and Mariama Drame and Quentin Lhoest and Alexander M. Rush",
    booktitle = "Proceedings of the 2020 Conference on Empirical Methods in Natural Language Processing: System Demonstrations",
    month = oct,
    year = "2020",
    address = "Online",
    publisher = "Association for Computational Linguistics",
    url = "https://www.aclweb.org/anthology/2020.emnlp-demos.6",
    pages = "38--45"
}

@article{dianjinocr,
  title={Dianjin-ocr-r1: Enhancing ocr capabilities via a reasoning-and-tool interleaved vision-language model},
  author={Chen, Qian and Zhang, Xianyin and Guo, Lifan and Chen, Feng and Zhang, Chi},
  journal={arXiv preprint arXiv:2508.13238},
  year={2025}
}
}

\appendix
\clearpage

\newcommand\DoToC{%
    \startcontents
    \printcontents{}{1}{\hrulefill\vskip0pt}
    \vskip0pt \noindent\hrulefill
    }

\setcounter{table}{0}
\setcounter{figure}{0}
\setcounter{equation}{0}
\setcounter{footnote}{0}
\renewcommand{\thetable}{S\arabic{table}}
\renewcommand{\thefigure}{S\arabic{figure}}
\renewcommand{\theequation}{S\arabic{equation}}
\renewcommand{\thealgorithm}{S\arabic{algorithm}}

\begin{center}
    \Large
    \textbf{Appendix}
    \vspace{1.0em}
\end{center}

\noindent\textbf{Overview}
\noindent\DoToC

\section{More Details of MPDocBench}
\subsection{Dataset Statistics}

\begin{figure}[t]
    \centering
\includegraphics[width=1.0\linewidth]{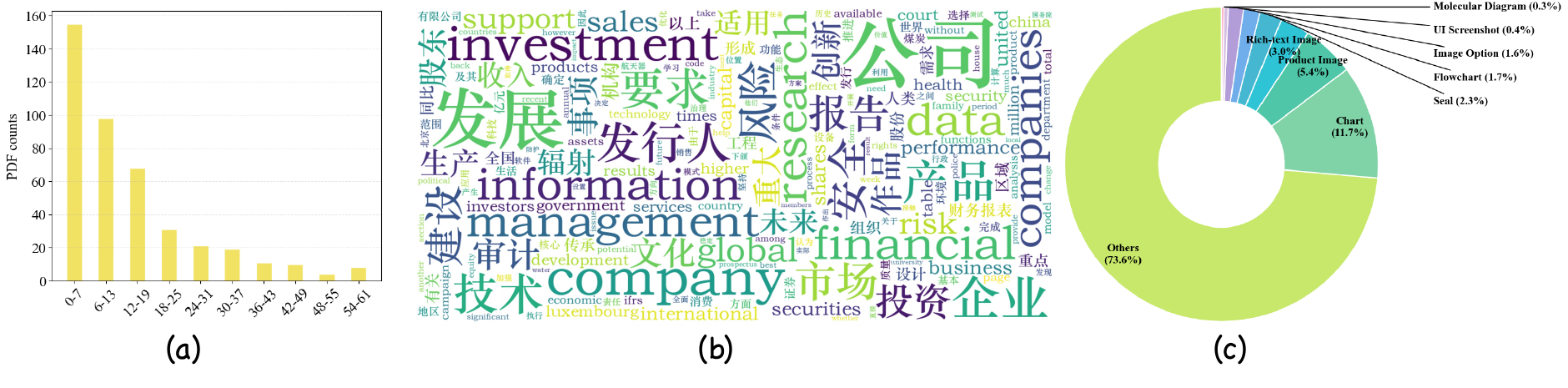}
   \caption{Visualization of the data distribution in MPDocBench-Parse. (a) Distribution of page number of documents. (b) Word cloud of document topics. (c) Distribution of figure subcategories.}\label{fig:supp_data_statistics}
\end{figure}

Due to space limitations in the main paper, we provide additional dataset statistics here to give a clearer picture of MPDocBench-Parse. These statistics further support our goal of building a practical benchmark for multi-page document parsing under realistic application settings.
As shown in Fig.~\ref{fig:supp_data_statistics}(a), the number of headings per PDF varies substantially, indicating that the dataset contains documents with both simple and dense heading organization. 
The topic visualization in Fig.~\ref{fig:supp_data_statistics}(b) further shows that MPDocBench-Parse is oriented toward realistic parsing scenarios. Unlike prior benchmarks such as READoc, which are more concentrated on academic-style documents, our benchmark covers broader real-world topics and application contexts.
Finally, Fig.~\ref{fig:supp_data_statistics}(c) shows the distribution of figure subcategories. In addition to the dominant others category, the dataset includes diverse visual content such as charts, product images, rich-text images, seals, flowcharts, UI screenshots, and molecular diagrams. This further demonstrates the visual diversity of MPDocBench-Parse and highlights its value for evaluating both content fidelity and logical structural correctness in practical document parsing.

\begin{table*}[h!]
\centering
\caption{Layout and Logic Structure Annotation Explanations and Statistics.}
\label{tab:structure_annotations}
\resizebox{1\textwidth}{!}{
\begin{tabular}{ l l l c }
\toprule
\multicolumn{4}{l}{\textbf{Layout Annotation}} \\
\hline
No. & Category Name & Explanation & Total \\
\hline
1 & Heading & Title, Section, subsection, or chapter titles within the main text. & 6438 \\
2 & Paragraph Heading & Minor titles or inline headings directly preceding a specific paragraph. & 18 \\
3 & TOC Heading & The title of the table of contents itself (e.g., "Contents"). & 182 \\
4 & Paragraph & Standard text blocks that form the main body of the document. & 28778 \\
5 & Table of Contents & The catalog section listing chapters, sections, and corresponding page numbers. & 277 \\
6 & Text Option & Text-based choices, commonly found in exams or questionnaires. & 826 \\
7 & Equation & Mathematical or scientific formulas, either inline or displayed. & 912 \\
8 & Code Block & Visual snippets or screenshots of programming source code. & 2 \\
9 & Table & Data or content organized in a structured row and column format. & 951 \\
10 & Table Caption & Descriptive text or title, typically placed above the table. & 344 \\
11 & Table Footnotes & Explanatory notes or remarks positioned directly below the table. & 320 \\
12 & Chart & Data visualizations such as bar charts, line graphs, and pie charts. & 350 \\
13 & Seal & Official stamps, seals, or watermarks superimposed on the document. & 69 \\
14 & Image Option & Image-based choices, used in visual questionnaires or tests. & 47 \\
15 & Rich-text Image & Complex graphic elements containing highly stylized or artistic text. & 90 \\
16 & Flowchart & Diagrams representing workflows, processes, or algorithms. & 51 \\
17 & Molecular Image & Visual representations of chemical molecules or compounds. & 8 \\
18 & Product Image & Photographs or illustrations specifically showcasing commercial products. & 162 \\
19 & UI Screenshot & Images capturing software user interfaces, apps, or web pages. & 13 \\
20 & Other Image & Various uncategorized images, illustrations, or graphical elements. & 2202 \\
21 & Image & General pictures or photographic content within the document. & 2 \\
22 & Figure Caption & Descriptive text or title, typically placed below the figure. & 636 \\
23 & Figure Footnotes & Additional textual explanations or remarks associated with a figure. & 981 \\
24 & Header & Information located at the top margin, often repeating across pages. & 2267 \\
25 & Header Image & Logos or graphical elements located specifically within the header area. & 368 \\
26 & Footer & Information located at the bottom margin, separate from the main content. & 1245 \\
27 & Footer Image & Logos or graphical elements located specifically within the footer area. & 330 \\
28 & Page Number & Numeric or Roman numeral indicators showing the page sequence. & 2282 \\
29 & Aside Text & Content placed in the side margins, separate from the main text flow. & 71 \\
30 & Footnote & Notes positioned at the bottom of the page providing extra context. & 157 \\
\hline
\multicolumn{4}{l}{\textbf{Logic Structure Annotation}} \\
\hline
No. & Relation Name & Explanation & Total \\
\hline
1 & Parent\_son & Represents hierarchical associations within the document, \ie heading-to-heading, heading-to-paragraph, or figure-to-caption relationships. & 43614 \\
2 & Truncated Text & Indicates a truncation relationship between two text blocks (\eg across pages or columns), implying they should be merged semantically. & 1577 \\
3 & Truncated Table & Indicates a truncation relationship between two table blocks, implying they belong to the same logical table and require merging. & 118 \\
\bottomrule
\end{tabular}
}
\end{table*}

\subsection{Annotation Statistics}
In the structural annotation stage, annotators are required to mark the location of each layout element using bounding boxes or polygons and assign a layout category to each annotated region. As summarized in Tab.~\ref{tab:structure_annotations}, our annotation scheme is highly fine-grained, covering 30 layout categories spanning textual, tabular, visual, and marginal content. Such detailed categorization enables a more comprehensive evaluation of document parsing systems, since models are required not only to detect coarse layout regions, but also to distinguish semantically and functionally different blocks.

In addition to region-level annotations, we further annotate structural relations between pairs of related layout blocks. Specifically, the relation types include parent\_son, which describes hierarchical dependencies between layout blocks, as well as truncated text and truncated table, which indicate that two text blocks or table blocks belong to the same logical content and should be merged. 
These relation annotations provide the basis for evaluating higher-level document parsing capabilities.
In particular, the explicit annotation of block relations enables us to recover heading hierarchies and identify text or table blocks that should be merged during evaluation. Therefore, the annotation design supports not only fine-grained layout analysis, but also the assessment of logical structural correctness and semantic continuity in realistic multi-page document parsing scenarios.

\subsection{Details about the Evaluation Metrics}\label{evaluation_detials}
\textbf{Figure Extraction.} 
For the evaluation of figure extraction, a key challenge is that different parsing systems may produce bounding boxes under different coordinates.
In particular, some specialized parsing systems take PDF documents as input, and different models may adopt different DPI settings when rendering PDF pages into images. As a result, even for the same document page, the image size may vary across systems, which in turn leads to different absolute bounding box coordinates.
To enable fair and consistent evaluation, we normalize all figure bounding boxes to a unified coordinate space ranging from 0 to 1000. For specialized parsing models, we first extract the predicted figure bounding boxes and then rescale them according to the size of the corresponding input image so that all coordinates are mapped into the same normalized range. In this way, the final bounding box representation becomes comparable across models, regardless of the original image resolution used during parsing. For general-purpose VLMs, we instead instruct the models through prompting to directly output normalized bounding boxes in the 0--1000 coordinate space.

\begin{algorithm}[htbp]
\caption{Heading Hierarchy Tree Construction and Evaluation}
\label{alg:head_teds}
\begin{algorithmic}[1]
\Require 
    Predicted heading list $\mathcal{P}$, a sequence of Markdown formatted strings where each element represents an extracted heading (e.g., ``\#\# Introduction''). Ground truth $\mathcal{G} = \langle R, N \rangle$, where $R$ is a list of structural edges (relations) between headings, and $N$ is a dictionary of node attributes (including heading content and reading order).

\Function{Evaluate}{$\mathcal{P}, \mathcal{G}$}
    \State $T_{pred} \gets$ \Call{BuildMDTree}{$\mathcal{P}$}
    \State $R, N \gets \mathcal{G}[0], \mathcal{G}[1]$
    \State $T_{gold} \gets$ \Call{BuildJSONTree}{$R, N$}
    \State $D \gets$ \Call{APTED}{$T_{pred}, T_{gold}$} \Comment{Compute Tree Edit Distance}
    \State $S \gets 1.0 - \frac{D}{\max(|T_{pred}|, |T_{gold}|)}$ \Comment{$|T|$ is the total number of nodes}
    \State \Return $S$
\EndFunction

\Statex
\Function{BuildMDTree}{$\mathcal{P}$}
    \State $root \gets$ \text{Node}(content=``root'', depth=0)
    \State $Stack \gets [root]$
    \For{each heading string $s \in \mathcal{P}$}
        \State $level \gets$ Count the number of '\#' in $s$
        \State $title \gets$ Strip '\#' and whitespaces from $s$
        \While{$level \le Stack.\text{top}().\text{depth}$}
            \State $Stack.\text{pop}()$ \Comment{Trace back to find the correct parent}
        \EndWhile
        \State $v \gets$ \text{Node}(content=$title$, depth=$level$)
        \State $v.parent \gets Stack.\text{top}()$
        \State $Stack.\text{top}().children.\text{append}(v)$
        \State $Stack.\text{push}(v)$
    \EndFor
    \State \Return $root$
\EndFunction

\Statex
\Function{BuildJSONTree}{$R, N$}
    \State $root \gets$ \text{Node}(content=``root'')
    \State $\mathcal{M} \gets \{ \text{``root''} : root \}$ \Comment{Initialize node map}
    
    \For{each relation $(u_{id}, v_{id}) \in R$}
        \If{$u_{id} \notin \mathcal{M}$}
            \State $\mathcal{M}[u_{id}] \gets$ \text{Node}(content=$N[u_{id}].\text{text}$, order=$N[u_{id}].\text{order}$)
        \EndIf
        \If{$v_{id} \notin \mathcal{M}$}
            \State $\mathcal{M}[v_{id}] \gets$ \text{Node}(content=$N[v_{id}].\text{text}$, order=$N[v_{id}].\text{order}$)
        \EndIf
        \State $\mathcal{M}[u_{id}].children.\text{append}(\mathcal{M}[v_{id}])$
        \State $\mathcal{M}[v_{id}].parent \gets \mathcal{M}[u_{id}]$
    \EndFor
    
    \For{each node $v \in \mathcal{M}.\text{values}() \setminus \{root\}$}
        \If{$v.parent = \text{None}$} \Comment{Attach orphan top-level nodes to root}
            \State $root.children.\text{append}(v)$
            \State $v.parent \gets root$
        \EndIf
    \EndFor
    
    \For{each node $v \in \mathcal{M}.\text{values}()$}
        \State Sort $v.children$ in ascending order based on node $order$
    \EndFor
    
    \State \Return $root$
\EndFunction

\end{algorithmic}
\end{algorithm}

\textbf{Heading Hierarchy.} 
For the evaluation of heading hierarchy recovery, we construct tree structures from the model-predicted heading sequence and the annotated ground-truth heading structure, and compare them based on tree edit distance. The overall procedure is summarized in Alg.~\ref{alg:head_teds}. For the predicted results, we parse the Markdown heading sequence produced by the model and determine the hierarchy level of each heading according to the number of `\#' symbols, thereby incrementally building the corresponding heading tree. For the ground truth, we construct a reference tree from the annotated heading nodes and their parent-child relations, and sort the child nodes according to their reading order so that the hierarchy is consistent with the logical organization of the document.
After obtaining the predicted tree and the ground-truth tree, we compute their tree edit distance using APTED and further normalize it into the final score. Specifically, the score is defined as $1-\frac{D}{\max(|T_{pred}|, |T_{gold}|)}$, where $D$ denotes the tree edit distance between the predicted tree and the ground-truth tree, and $|T_{pred}|$ and $|T_{gold}|$ denote the total numbers of nodes in the two trees, respectively. This metric jointly evaluates heading content, hierarchical relations, and overall structural organization, thereby providing a comprehensive assessment of model performance on heading hierarchy recovery.

\subsection{Details about OCR Label Refinement}\label{ocr_label_refinement}

As described in the main text, samples with $Sim_{avg} \ge 0.95$ indicate a high level of agreement among the three recognition models on a given text block. However, such agreement does not necessarily guarantee that the annotations are correct, since multiple models may still produce the same erroneous prediction. To further improve the annotation quality of these text blocks, we employ a large vision-language model (Gemini-3-Pro) as an adjudicator to further verify and refine the candidate annotations.
Specifically, we provide Gemini-3-Pro with the original text block image together with the three candidate recognition results, and instruct it to compare these candidates against the visual content of the image in order to generate the refined annotation. 
In this process, the model can not only leverage the agreement among the candidate texts, but also directly refer to character shapes, contextual cues, and local visual details in the original image to correct potential recognition errors.
Through this adjudication and refinement process, we obtain a more reliable final label for each text block sample. 
This strategy effectively reduces the noise caused by cases where multiple models make the same mistake, thereby substantially improving the overall quality and reliability of OCR annotations.

\subsection{Details about Human Annotators}
Our annotation pipeline comprises two stages, and separate groups of external annotators are therefore assigned to structure annotation and content annotation, respectively.

Specifically, for the structure annotation stage, we employ 10 annotators and 3 reviewers to complete the task. All annotators are between 25 and 35 years old, hold bachelor's degrees, and have prior experience in document parsing-related annotation, which helps ensure annotation quality. In addition, before the formal annotation process begins, we provide dedicated training to all annotators to further improve their understanding of the task definitions and annotation guidelines. Each structurally annotated document is visualized and then inspected by the 3 reviewers. A document is considered successfully annotated only if all three reviewers approve it.

For the content annotation stage, we employ 15 annotators and 3 reviewers. Their age and educational backgrounds are similar to those of the structure annotators, but they are more specialized in content-level annotation tasks, such as text block, table, and formula annotation. Due to the complexity of table and formula annotations, the reviewers verify annotation quality by comparing the rendered visualizations of the annotations against the original cropped images, thereby ensuring the accuracy and consistency of the final annotations.

\begin{figure}[!h]
    \centering
\includegraphics[width=0.91\linewidth]{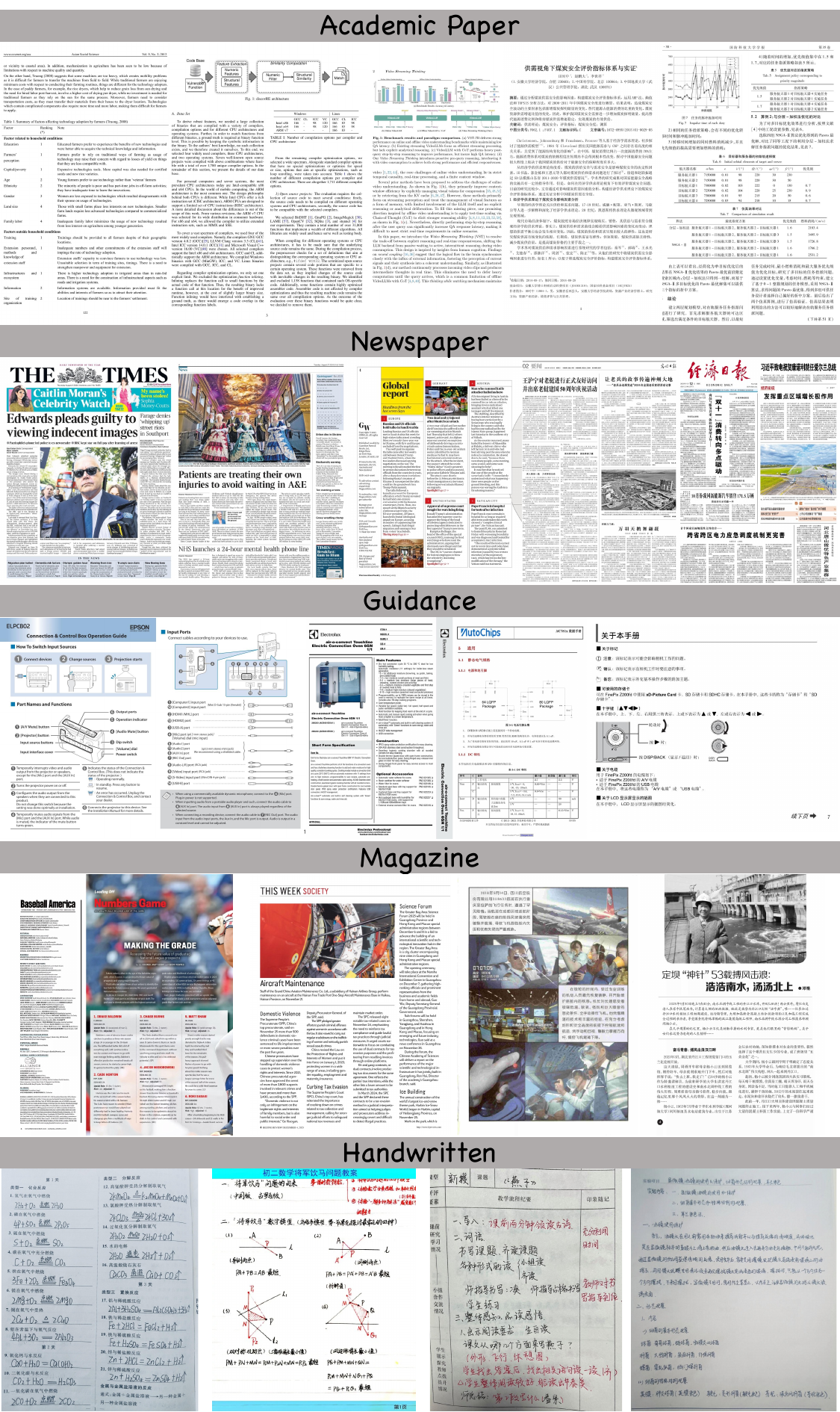}
   \caption{Some Examples in MPDocBench-Parse.}\label{fig:supp_benchsample1}
\end{figure}

\begin{figure}[!h]
    \centering
\includegraphics[width=0.91\linewidth]{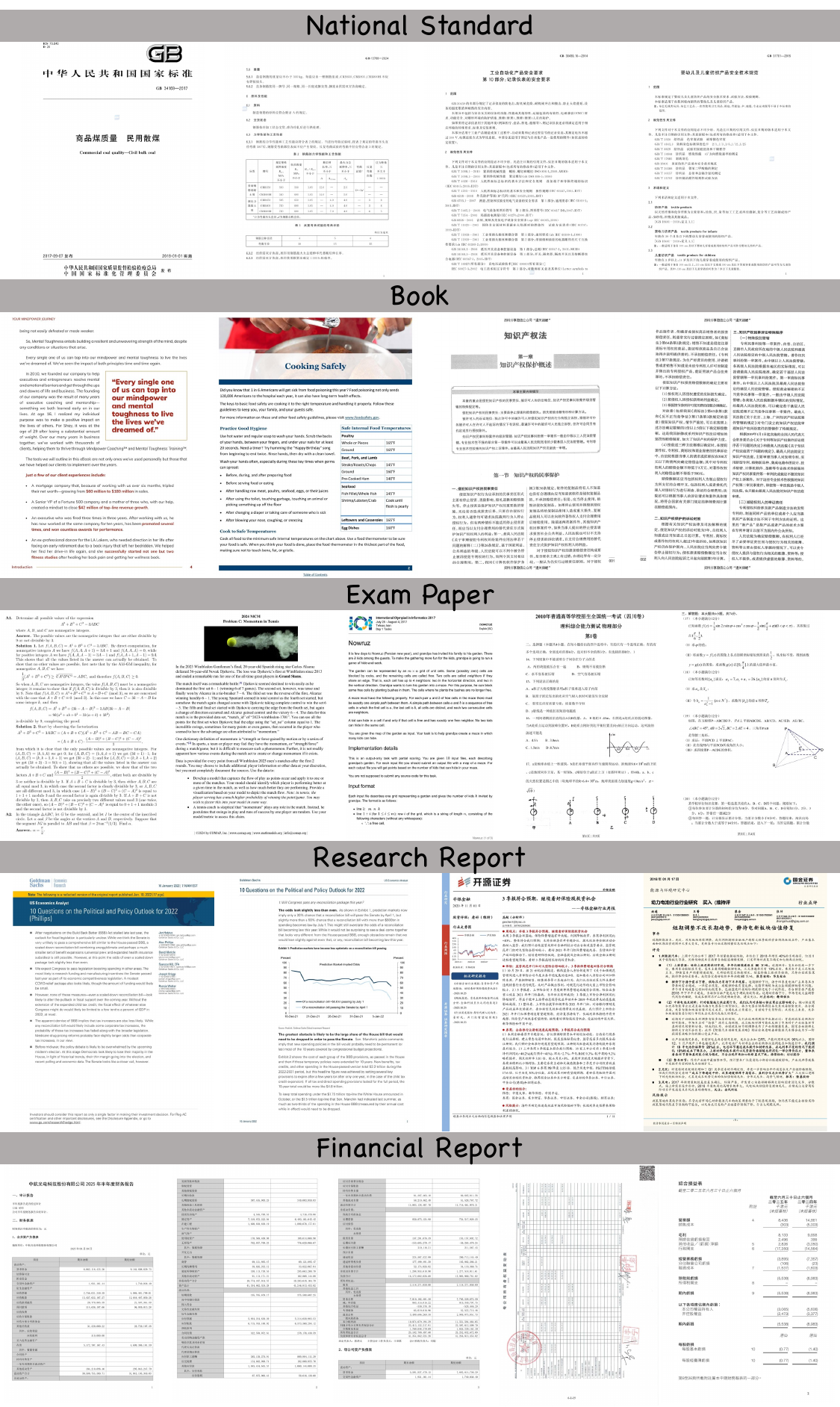}
   \caption{Some Examples in MPDocBench-Parse.}\label{fig:supp_benchsample2}
\end{figure}

\begin{figure}[!h]
    \centering
\includegraphics[width=0.92\linewidth]{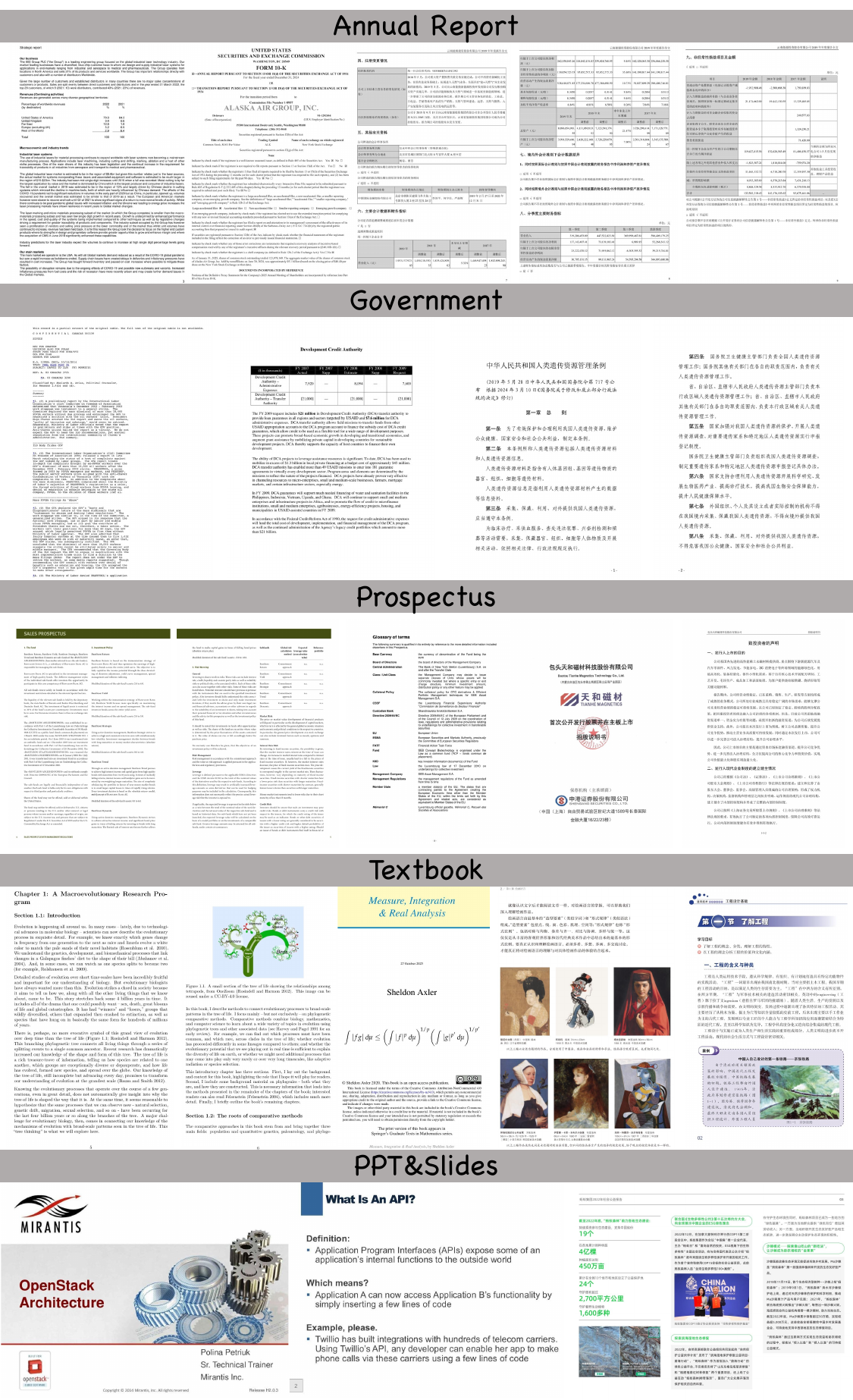}
   \caption{Some Examples in MPDocBench-Parse.}\label{fig:supp_benchsample3}
\end{figure}

\subsection{Benchmark Samples}\label{supp_benchmark_samples}
Figures~\ref{fig:supp_benchsample1}-\ref{fig:supp_benchsample3} present sample documents from each document category. It can be clearly observed that academic papers, government documents, exam papers, and national standards generally have relatively simple layouts. 
From a structural perspective, these documents are typically organized in either single-column or double-column formats, with mixed or multi-column layouts being rare. 
In addition, their reading order is usually regular and mostly follows a top-to-bottom pattern, with few complicated reading order.
Regarding heading hierarchy, these document types often distinguish headings from body text through bold fonts and use numbering to indicate different heading levels. Such explicit visual and textual cues help models more accurately recover the hierarchical structure of documents, which is consistent with the findings reported in the main paper.
The dataset also includes document types with much more complex layouts, such as magazines, newspapers, manuals, handwritten documents, PPT\&Slides, and various reports. These documents usually exhibit more complicated layout structures, more diverse reading orders, and less obvious visual distinction between headings and body text. Meanwhile, their heading structures are often more complex, \ie with deeper hierarchy trees. They also tend to contain more semantically continuous content spanning different regions and richer visual elements, which further increases the parsing difficulty. As a result, current models generally achieve less satisfactory performance on global heading hierarchy recovery for these document types.

\section{More Experimental Analysis}

\subsection{Implementation Details}
We conduct inference for all open-source models using $8\times80$G GPUs. Following the official implementations provided by each model, we deploy Youtu-Parsing with \texttt{transformers}~\cite{transformers}; GLM-OCR, PaddleOCR-VL-1.5, MinerU2.5, dots.mocr, FireRed-OCR, dots.ocr, DeepSeek-OCR2, Dolphin-V2, OCR-Verse, Logics-Parsing-V2, Qianfan-OCR, and ChandraOCR 2 with \texttt{vLLM}~\cite{vllm}; and InternVL-3.5-38B together with MonkeyOCR-pro-3B using \texttt{lmdeploy}~\cite{lmdeploy}. For the remaining closed-source models, including Gemini-3.1-pro-preview, ChatGPT-5.2, Qwen3.6-plus, and Qwen3-VL-235B-A22B-Instruct, we access them via API calls. During inference, the temperature is set to 0 for all models, and the maximum number of generated tokens is set to 16,384. The detailed allocation of computational resources is as follows.

\textbf{GLM-OCR and PaddleOCR-VL-1.5.} We allocate 7 GPUs to deploy the corresponding VLM-based recognition models, with one model instance on each GPU, while the remaining GPU is used to load the layout detection model required in the parsing pipeline.

\textbf{MinerU2.5, dots.mocr, dots.ocr, DeepSeek-OCR2, Youtu-Parsing, Qianfan-OCR, and MonkeyOCR-pro-3B.} We deploy one model instance on each GPU, resulting in 8 parallel model instances to improve inference efficiency.

\textbf{FireRed-OCR, Dolphin-V2, OCR-Verse, Logics-Parsing-V2, and ChandraOCR 2.} For these models, we use \texttt{vLLM} with \texttt{tensor-parallel-size=4}, and  deploy 2 model instances in total.

\textbf{InternVL-3.5-38B.} Given its relatively large parameter scale, we set \texttt{tensor-parallel-size=8} so that the model is deployed in a fully tensor-parallel manner across all 8 GPUs during inference.

\begin{figure}[!h]
    \centering
\includegraphics[width=1.0\linewidth]{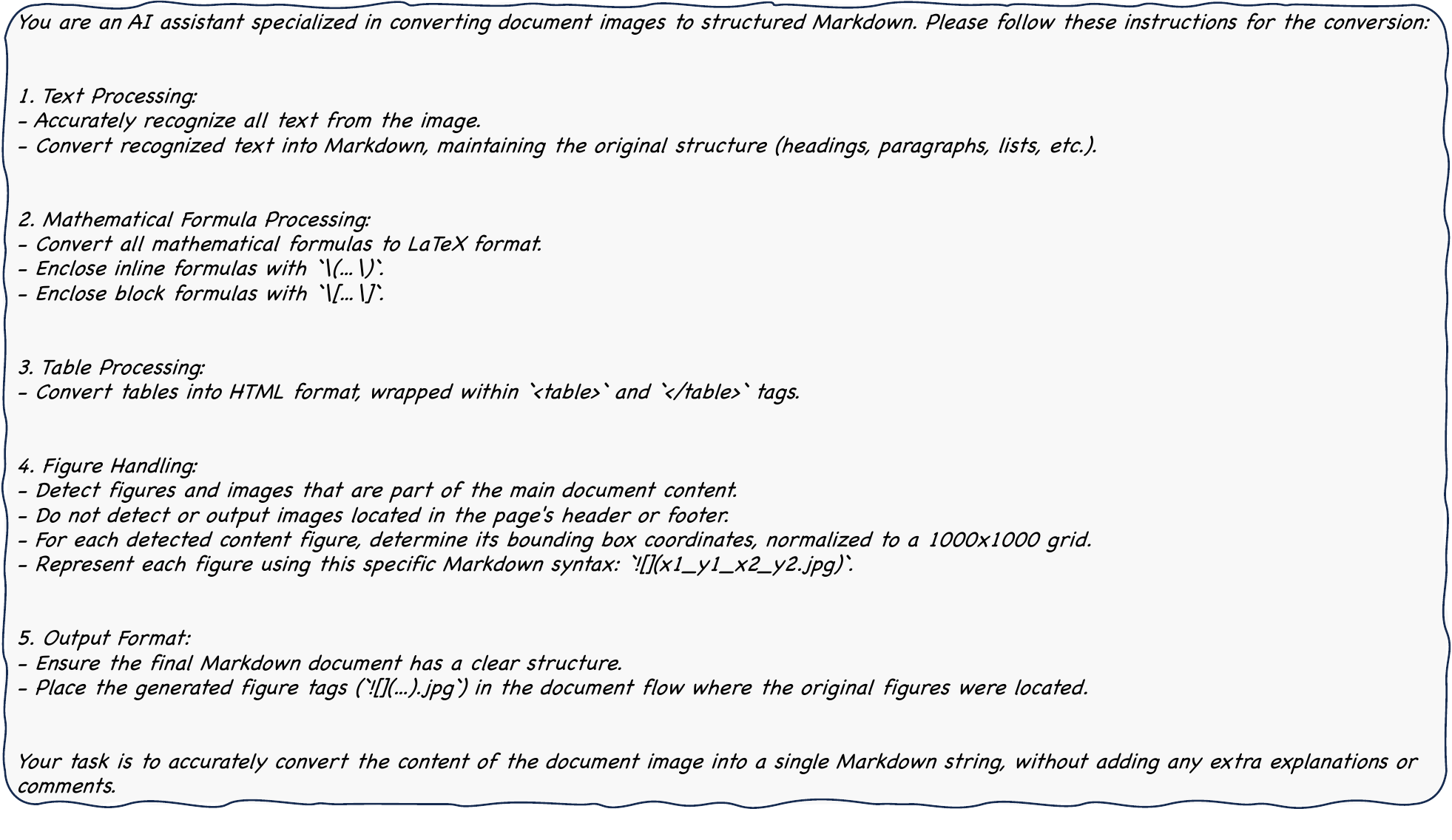}
    \vspace{-20pt}
   \caption{The parsing prompt for general VLMs during inference.}\label{fig:infer_prompt}
\end{figure}

\subsection{Inference Prompts for General VLMs}
Since specialized VLMs are intrinsically tailored for document parsing tasks and equipped with customized prompts, we retain their default prompts during inference without modification. These models are generally designed with task-specific architectures or optimization objectives, and their built-in prompts have already been tuned to match their parsing pipelines. Modifying such prompts may disrupt their original inference behavior and lead to suboptimal performance. Conversely, general VLMs necessitate carefully crafted instructions to effectively execute our designated parsing tasks, as they are not explicitly trained for fine-grained document parsing by default. Therefore, their prompts need to clearly specify the expected output format and emphasize the recognition of textual, structural, and visual elements.
Specifically, we adapt the prompt templates from OmniDocBench and MDPBench with targeted modifications. In addition to preserving the core instruction style of these prior benchmarks, we further refine the prompts to better align with our evaluation setting. The primary objective of these refinements is to guide the models to accurately recognize and localize image elements within the documents, while also improving their understanding of document structure and layout semantics. The detailed prompts are provided in Fig.~\ref{fig:infer_prompt}.

\begin{table*}[t]
  \centering
  \caption{Overall performance across different document types.}
\vspace{-7pt}
  \renewcommand{\arraystretch}{1.0}
  \resizebox{1.0\textwidth}{!}{
    \begin{tabular}{c| c c c c c c c c c c c c c c c}
    \toprule
    \makecell{\textbf{Model}} & 
    \makecell{\textbf{Annual}\\\textbf{Report}} & 
    \makecell{\textbf{Book}} &
    \makecell{\textbf{Financial}\\\textbf{Report}} &
    \makecell{\textbf{Exam}\\\textbf{Paper}} &
    \makecell{\textbf{Government}} & 
    \makecell{\textbf{National}\\\textbf{Standard}} & 
    \makecell{\textbf{Handwritten}} &
    \makecell{\textbf{Magazine}} & 
    \makecell{\textbf{Guidance}} &
    \makecell{\textbf{Newspaper}} &   
    \makecell{\textbf{Academic}\\\textbf{Paper}} &
    \makecell{\textbf{PPT\&}\\\textbf{Slides}} &
    \makecell{\textbf{Textbook}} &
    \makecell{\textbf{Research}\\\textbf{Report}} & 
    \makecell{\textbf{Prospectus}} \\
    \midrule
\multicolumn{16}{l}{\textbf{\textit{Pipeline-based Specialized VLMs}}} \\
\midrule
GLM-OCR~\cite{glmocr} & 77.84 & 67.30 & 68.84 & 85.05 & 74.08 & \textbf{78.34} & 68.39 & 76.88 & 75.99 & 75.19 & 82.13 & 74.76 & 72.84 & 77.76 & 69.28\\

PaddleOCR-VL-1.5~\cite{paddleocrvl15} & \textbf{80.71} & 79.05 & \textbf{83.94} & 86.88 & 76.93 & 70.95 & 68.29 & \textbf{79.52} & \textbf{77.97} & \textbf{80.22} & \textbf{87.22} & 77.52 & 73.12 & 77.21 & \textbf{76.45}\\

MinerU2.5~\cite{mineru25} & 80.12 & 66.84 & 75.78 & 84.26 & \textbf{81.53} & 66.31 & 63.21 & 72.65 & 73.64 & 74.11 & 83.50 & 74.36 & 69.95 & \textbf{82.93} & 70.51\\

Youtu-Parsing~\cite{youtuparsing} & 78.11 & 74.00 & 71.56 & 86.08 & 80.46 & 73.63 & 68.42 & 76.23 & 76.87 & 66.21 & 76.45 & 75.33 & 70.66 & 71.86 & 70.14\\

MonkeyOCR-pro-3B~\cite{monkeyocr} & 77.48 & 74.87 & 65.19 & 86.67 & 74.37 & 67.05 & 66.83 & 76.02 & 77.40 & 72.27 & 80.68 & 73.49 & 71.00 & 77.71 & 68.49\\

Dolphin-v2~\cite{dolphinv2} & 75.81 & 78.47 & 65.81 & 83.58 & 79.34 & 74.78 & 51.34 & 69.99 & 75.81 & 74.09 & 77.93 & 69.88 & 69.47 & 75.85 & 71.26\\

\midrule
\multicolumn{16}{l}{\textbf{\textit{End-to-End Specialized VLMs}}} \\
\midrule

dots.mocr~\cite{dotsmocr} & 75.52 & 77.81 & 73.61 & 87.12 & 76.21 & 71.89 & 50.87 & 76.63 & 74.26 & 69.35 & 77.98 & 64.91 & 68.95 & 75.42 & 67.46\\

FireRed-OCR~\cite{fireredocr} & 71.29 & 72.47 & 66.93 & 77.13 & 68.64 & 63.49 & 61.70 & 60.97 & 68.74 & 68.17 & 73.05 & 64.42 & 63.64 & 68.54 & 67.07\\

dots.ocr~\cite{dotsocr} & 76.00 & 73.23 & 73.75 & \textbf{87.26} & 78.03 & 63.64 & 54.22 & 68.30 & 74.70 & 72.34 & 78.42 & 76.81 & 70.05 & 77.61 & 69.28\\

DeepSeek-OCR2~\cite{deepseekocr2} & 77.02 & 67.29 & 74.43 & 86.28 & 78.77 & 72.52 & 61.48 & 74.22 & 76.25 & 70.58 & 83.41 & 75.98 & \textbf{73.52} & 80.56 & 72.79\\

OCRVerse~\cite{ocrverse} & 67.34 & 68.93 & 64.27 & 74.62 & 64.51 & 61.28 & 63.43 & 67.54 & 62.44 & 59.72 & 68.76 & 68.98 & 62.29 & 68.12 & 62.88\\

Logics-Parsing-v2~\cite{logicsparsing} & 76.29 & 78.47 & 70.35 & 86.29 & 79.23 & 68.10 & 69.37 & 74.54 & 77.25 & 71.40 & 78.30 & \textbf{82.23} & 72.48 & 79.74 & 65.93\\

Qianfan-OCR~\cite{qianfanocr} & 78.78 & \textbf{79.08} & 69.30 & 86.65 & 76.42 & 76.44 & 67.14 & 78.70 & 77.89 & 49.32 & 76.82 & 80.47 & 70.48 & 72.86 & 74.60\\

ChandraOCR 2~\cite{chandraocr2} & 79.08 & 78.38 & 71.00 & 83.79 & 80.71 & 65.51 & \textbf{71.61} & 79.44 & 75.21 & 73.12 & 75.69 & 73.92 & 69.38 & 80.76 & 69.81\\

\midrule
\multicolumn{16}{l}{\textbf{\textit{General VLMs}}} \\
\midrule

Gemini-3.1-pro-preview~\cite{gemini31} & 76.60 & 74.58 & 65.23 & 80.75 & 71.44 & 62.05 & 64.86 & 73.49 & 72.65 & 72.20 & 73.16 & 77.74 & 67.86 & 70.72 & 68.43\\

ChatGPT-5.2-2025-12-11~\cite{chatgpt} & 69.87 & 74.17 & 64.05 & 73.17 & 68.49 & 64.79 & 55.95 & 70.13 & 68.80 & 46.90 & 71.81 & 74.09 & 67.94 & 67.84 & 62.08\\

Qwen3.6-plus~\cite{bailian} & 74.99 & 73.93 & 62.09 & 79.42 & 72.01 & 73.67 & 60.98 & 75.81 & 68.99 & 62.33 & 72.51 & 80.52 & 67.14 & 72.24 & 65.89\\

Qwen3-VL-235B~\cite{qwen3vl} & 80.38 & 78.45 & 68.72 & 82.78 & 78.39 & 75.16 & 62.99 & 78.87 & 72.68 & 69.39 & 77.04 & 77.71 & 70.76 & 72.49 & 69.05\\

InternVL-3.5-38B~\cite{internvl35} & 64.79 & 60.33 & 62.04 & 73.17 & 65.22 & 58.05 & 54.56 & 61.07 & 59.40 & 36.47 & 62.60 & 69.32 & 62.37 & 60.07 & 58.46\\

\bottomrule
\end{tabular}%
}
\label{tab:supp_document_type_score}
\vspace{-15pt}
\end{table*}

\begin{table*}[t]
  \centering
  \caption{Table recognition performance (TEDS) across different document types.}
\vspace{-7pt}
  \renewcommand{\arraystretch}{1.0}
  \resizebox{1.0\textwidth}{!}{
    \begin{tabular}{c| c c c c c c c c c c c c c c c}
    \toprule
    \makecell{\textbf{Model}} & 
    \makecell{\textbf{Annual}\\\textbf{Report}} & 
    \makecell{\textbf{Book}} &
    \makecell{\textbf{Financial}\\\textbf{Report}} &
    \makecell{\textbf{Exam}\\\textbf{Paper}} &
    \makecell{\textbf{Government}} & 
    \makecell{\textbf{National}\\\textbf{Standard}} & 
    \makecell{\textbf{Handwritten}} &
    \makecell{\textbf{Magazine}} & 
    \makecell{\textbf{Guidance}} &
    \makecell{\textbf{Newspaper}} &   
    \makecell{\textbf{Academic}\\\textbf{Paper}} &
    \makecell{\textbf{PPT\&}\\\textbf{Slides}} &
    \makecell{\textbf{Textbook}} &
    \makecell{\textbf{Research}\\\textbf{Report}} & 
    \makecell{\textbf{Prospectus}} \\
    \midrule
\multicolumn{16}{l}{\textbf{\textit{Pipeline-based Specialized VLMs}}} \\
\midrule

GLM-OCR~\cite{glmocr} & 87.18 & 87.89 & 80.99 & 93.62 & 66.14 & 87.29 & 43.70 & 78.46 & 79.69 & 52.50 & 85.21 & 67.64 & 43.19 & 89.32 & 81.94\\

PaddleOCR-VL-1.5~\cite{paddleocrvl15} & 89.15 & 86.25 & 91.31 & 93.28 & 63.59 & 78.84 & 40.30 & \textbf{83.98} & 73.74 & 53.33 & 86.03 & 50.44 & 39.21 & 87.03 & 85.40\\

MinerU2.5~\cite{mineru25} & 90.72 & 87.65 & \textbf{91.99} & 86.32 & 67.12 & 83.21 & 40.48 & 42.99 & 80.52 & 53.33 & 85.86 & 90.09 & 38.44 & 80.47 & \textbf{86.46}\\

Youtu-Parsing~\cite{youtuparsing} & 89.12 & 81.25 & 82.40 & 88.46 & 71.76 & \textbf{91.36} & 55.94 & 61.50 & 78.10 & 7.67 & \textbf{87.97} & 63.98 & 29.37 & 87.22 & 79.27\\

MonkeyOCR-pro-3B~\cite{monkeyocr} & 85.86 & 87.63 & 66.30 & 94.68 & 58.02 & 80.17 & 45.53 & 63.36 & 76.86 & 44.44 & 83.94 & 57.79 & 40.72 & 82.97 & 79.07\\

Dolphin-v2~\cite{dolphinv2} & 87.62 & 80.38 & 77.67 & 94.65 & 64.05 & 88.73 & 41.95 & 26.49 & 84.11 & 48.44 & 86.93 & 49.99 & 26.18 & 84.84 & 81.67\\

\midrule
\multicolumn{16}{l}{\textbf{\textit{End-to-End Specialized VLMs}}} \\
\midrule

dots.mocr~\cite{dotsmocr} & 90.68 & 79.34 & 80.39 & 88.55 & 62.78 & 84.26 & 16.22 & 73.39 & 74.59 & 50.00 & 86.29 & 27.68 & 28.28 & 85.38 & 77.72\\

FireRed-OCR~\cite{fireredocr} & 87.66 & 80.41 & 78.26 & 88.39 & 55.41 & 86.27 & 61.05 & 0.00 & 76.31 & 45.00 & 85.31 & 57.76 & 29.47 & 84.81 & 75.65\\

dots.ocr~\cite{dotsocr} & 89.01 & 78.94 & 80.21 & 88.76 & 62.82 & 83.52 & 36.70 & 0.00 & 75.07 & \textbf{58.00} & 85.86 & 72.04 & 31.60 & 87.32 & 82.92\\

DeepSeek-OCR2~\cite{deepseekocr2} & 86.33 & 88.21 & 79.52 & 86.76 & 62.67 & 84.97 & 33.70 & 49.48 & 75.98 & 50.00 & 84.36 & 44.13 & 28.19 & 88.74 & 71.07\\

OCRVerse~\cite{ocrverse} & 90.55 & 80.87 & 77.88 & \textbf{96.47} & 63.69 & 83.80 & 66.56 & 73.74 & 78.48 & 41.67 & 87.28 & 92.73 & 40.87 & 85.61 & 78.03\\

Logics-Parsing-v2~\cite{logicsparsing} & 87.92 & 86.70 & 78.82 & 89.02 & 71.01 & 82.14 & 67.56 & 49.81 & 79.86 & 48.44 & 83.64 & \textbf{94.56} & 37.84 & 89.67 & 73.61\\

Qianfan-OCR~\cite{qianfanocr} & 85.00 & \textbf{89.32} & 76.37 & 95.32 & 62.84 & 86.27 & 48.42 & 79.17 & 75.98 & 45.83 & 84.55 & 84.63 & 41.27 & 84.51 & 74.64\\

ChandraOCR 2~\cite{chandraocr2} & \textbf{91.10} & 87.90 & 79.53 & 86.76 & \textbf{75.37} & 84.16 & 61.75 & 77.86 & \textbf{86.25} & 57.14 & 85.12 & 73.18 & 29.62 & \textbf{91.20} & 83.09\\

\midrule
\multicolumn{16}{l}{\textbf{\textit{General VLMs}}} \\
\midrule

Gemini-3.1-pro-preview~\cite{gemini31} & 89.83 & 81.69 & 81.39 & 81.71 & 58.25 & 74.75 & 63.77 & 44.34 & 81.00 & 50.00 & 86.92 & 71.58 & 16.70 & 87.14 & 75.34\\

ChatGPT-5.2-2025-12-11~\cite{chatgpt} & 88.59 & 79.06 & 75.74 & 87.26 & 60.08 & 71.85 & 57.29 & 47.45 & 72.37 & 1.56 & 82.48 & 63.10 & 35.10 & 79.96 & 69.67\\

Qwen3.6-plus~\cite{bailian} & 89.83 & 82.64 & 79.92 & 81.30 & 62.79 & 83.88 & 64.51 & 79.16 & 77.71 & 0.00 & 86.13 & 78.21 & 28.94 & 85.84 & 65.36\\

Qwen3-VL-235B~\cite{qwen3vl} & 89.76 & 89.10 & 79.10 & 88.84 & 66.83 & 87.59 & \textbf{70.88} & 76.46 & 75.97 & 38.89 & 81.92 & 56.20 & \textbf{46.41} & 79.85 & 83.71\\

InternVL-3.5-38B~\cite{internvl35} & 81.48 & 68.39 & 66.41 & 83.97 & 57.72 & 66.30 & 55.59 & 35.22 & 63.58 & 0.00 & 75.78 & 59.64 & 23.42 & 69.83 & 70.28\\

\bottomrule
\end{tabular}%
}
\label{tab:supp_table_teds_across_document_types}
\vspace{-15pt}
\end{table*}

\subsection{More Experimental Results}
\textbf{Overall performance across different document types.} Tab.~\ref{tab:supp_document_type_score} reports the overall performance of different models across document types. The results show clear variations across categories, suggesting that document parsing difficulty depends not only on model capability, but also on layout complexity, visual composition, and document structure.

Overall, models perform relatively well on exam paper, academic paper, annual report, and financial report. These document types usually have more regular layouts, clearer structural organization, or more explicit semantic relationships, making them easier to parse. 
In particular, exam and academic paper appear more compatible with current models, likely because of their relatively stable page structures and stronger template characteristics.
By contrast, performance is weaker on handwritten, PPT\&Slides, textbook, prospectus, and newspaper. These document types often involve greater visual complexity or more challenging layouts. For example, Handwritten documents suffer from diverse writing styles and blurry character boundaries, while PPT\&Slides and newspaper often contain irregular layouts, strong visual interference, or dense text-image compositions. Textbooks also remain difficult because they frequently include figures, formulas, annotation boxes, and cross-region layouts. 
Further observation shows that the relative strengths of different models are not entirely consistent across document types, suggesting that current document parsing capability remains highly scenario-dependent. 
This also indicates that existing methods can more easily learn effective patterns from structurally stable pages, while still showing clear limitations in scenarios with complex visual layouts and weak structural priors.

\textbf{Table recognition performance across different document types.}
As shown in Tab.~\ref{tab:supp_table_teds_across_document_types}, table recognition performance varies noticeably across document types. This variation is broadly related to the layout structure, table distribution patterns, and visual complexity of the documents. 
For document types with clearer table boundaries, explicit row and column separators, and more stable contextual organization, such as reports, exam papers, and PPT\&Slides, models generally achieve higher TEDS scores, suggesting that current methods are more effective in relatively regular table scenarios. 
In contrast, performance is weaker on handwritten, newspaper, and textbook documents. 
Handwritten documents pose greater challenges due to diverse writing styles and blurry character boundaries, while newspapers and textbooks more often contain borderless tables, which makes both table localization and structure recovery more difficult. Further inspection of the predictions shows that tables in these document types are also more likely to be missed and eventually misclassified as plain text blocks.

\subsection{Visualization of Inference Results.}\label{visualization_inference}
\begin{figure}[t]
    \centering
\includegraphics[width=1.0\linewidth]{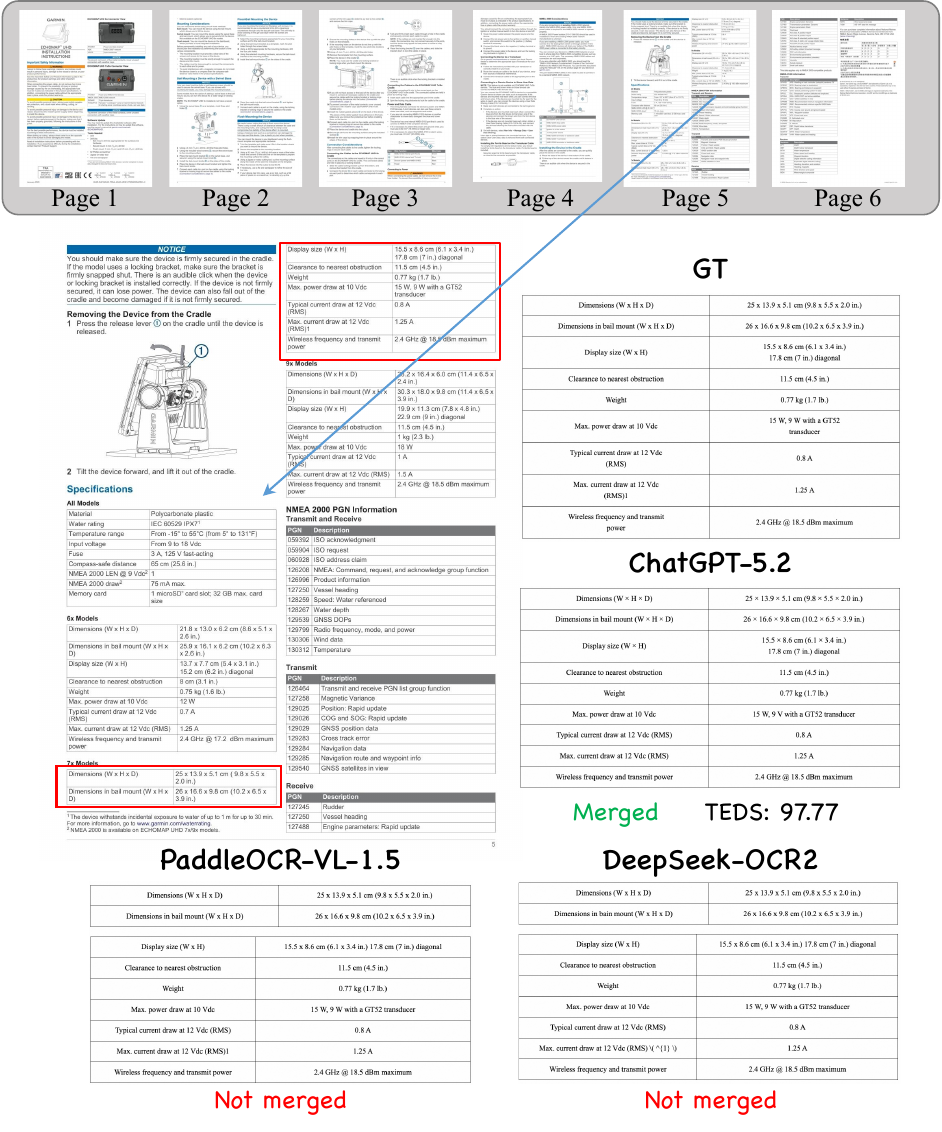}
    \vspace{-20pt}
   \caption{Examples of within-page truncated table merging.}\label{fig:supp_within_table_merge_success}
\end{figure}

\begin{figure}[!h]
    \centering
\includegraphics[width=1.0\linewidth]{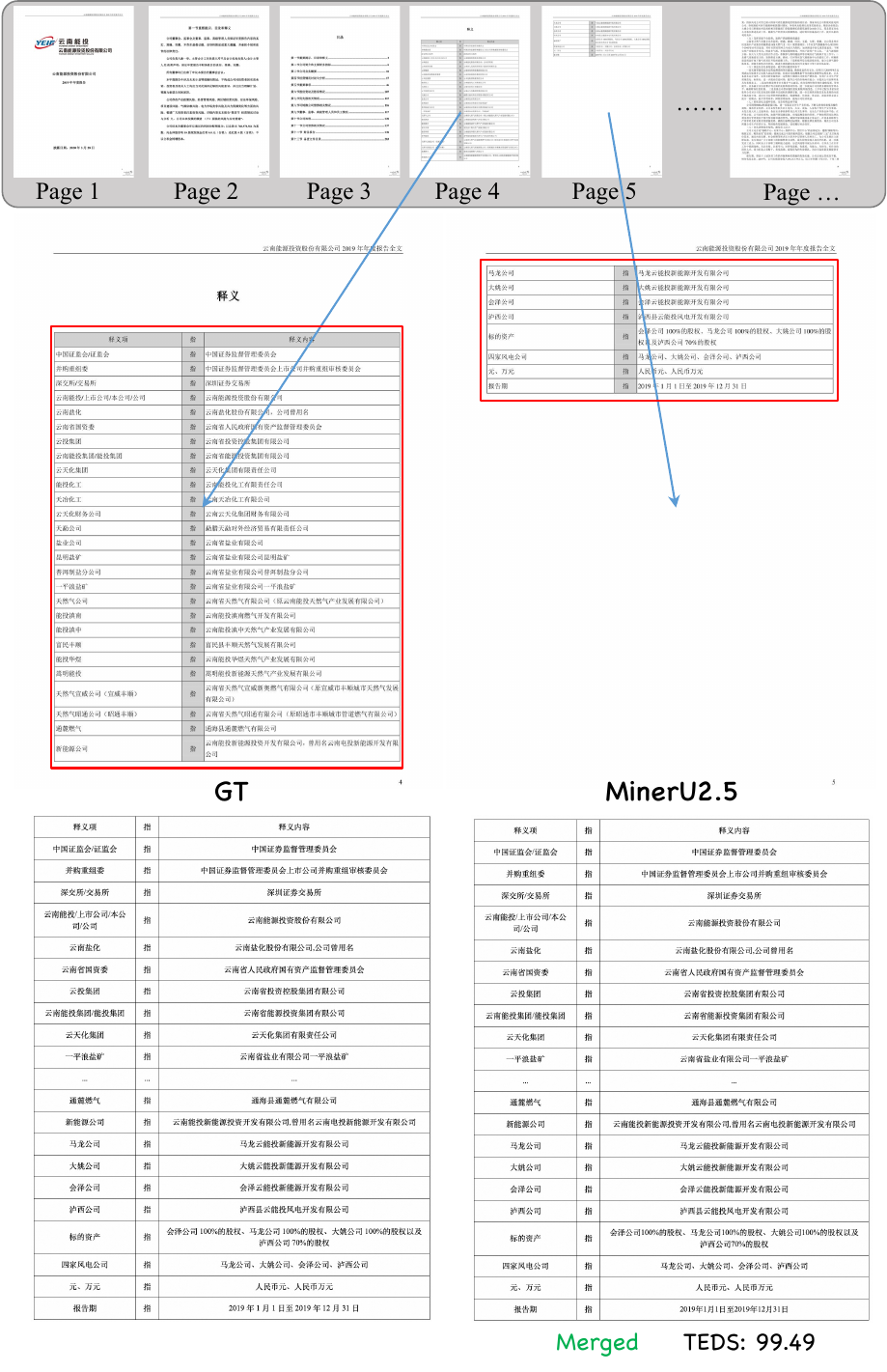}
    \vspace{-20pt}
   \caption{Examples of cross-page truncated table merging.}\label{fig:supp_cross_table_merge_success}
\end{figure}

\begin{figure}[!h]
    \centering
\includegraphics[width=0.95\linewidth]{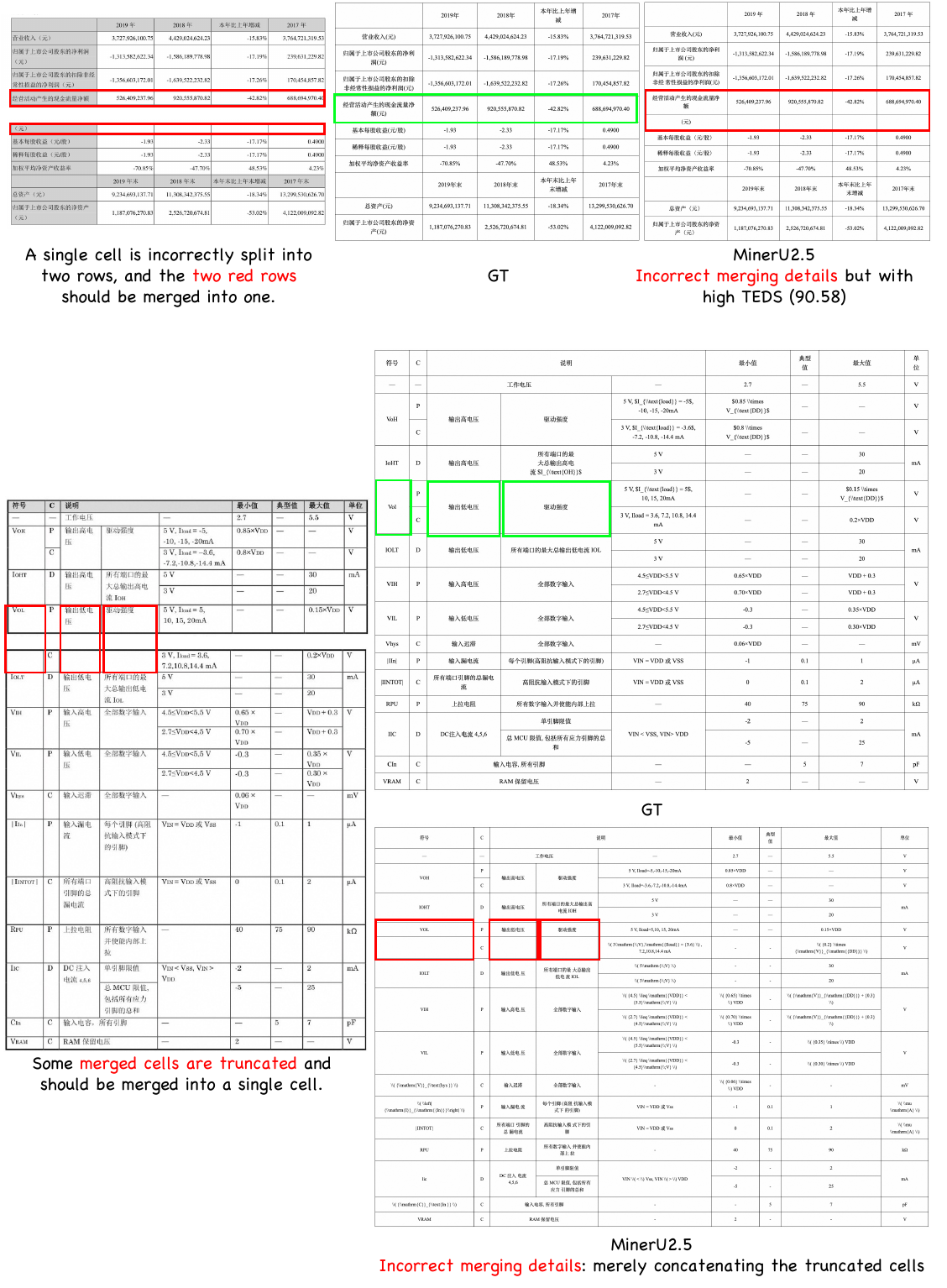}
   \caption{Visualization of truncated table merging details.}\label{fig:supp_table_merge}
\end{figure}

\textbf{Truncated Table Merging.} Figures~\ref{fig:supp_within_table_merge_success}-\ref{fig:supp_table_merge} present several cases of truncated table merging. Fig.~\ref{fig:supp_within_table_merge_success} and Fig.~\ref{fig:supp_cross_table_merge_success} show successful cases of within-page and cross-page merging, respectively. These examples suggest that the main challenge is to determine, based on table semantics, whether two spatially adjacent tables belong to the same truncated table and should be merged.
Notably, once a model correctly identifies two tables as belonging to the same truncated table, the final TEDS score can remain relatively high even when the merging details are not fully correct. This indicates that future evaluation should consider not only whether merging is performed, but also whether the merging details are structurally correct.
Fig.~\ref{fig:supp_table_merge} further illustrates several complex cases. First, an originally complete cell may be split into two rows during pagination, requiring the model to decide whether the text content should be concatenated and restored as a single cell. Second, original row-span cells may be truncated into upper and lower parts. If the model simply concatenates these parts without restoring the merged cell region, the resulting table structure remains incorrect, although its TEDS score is often still higher than that of not merging. In addition, some truncated table groups contain repeated headers, which should be removed before merging to obtain the correct final result.

\begin{figure}[!h]
    \centering
\includegraphics[width=0.95\linewidth]{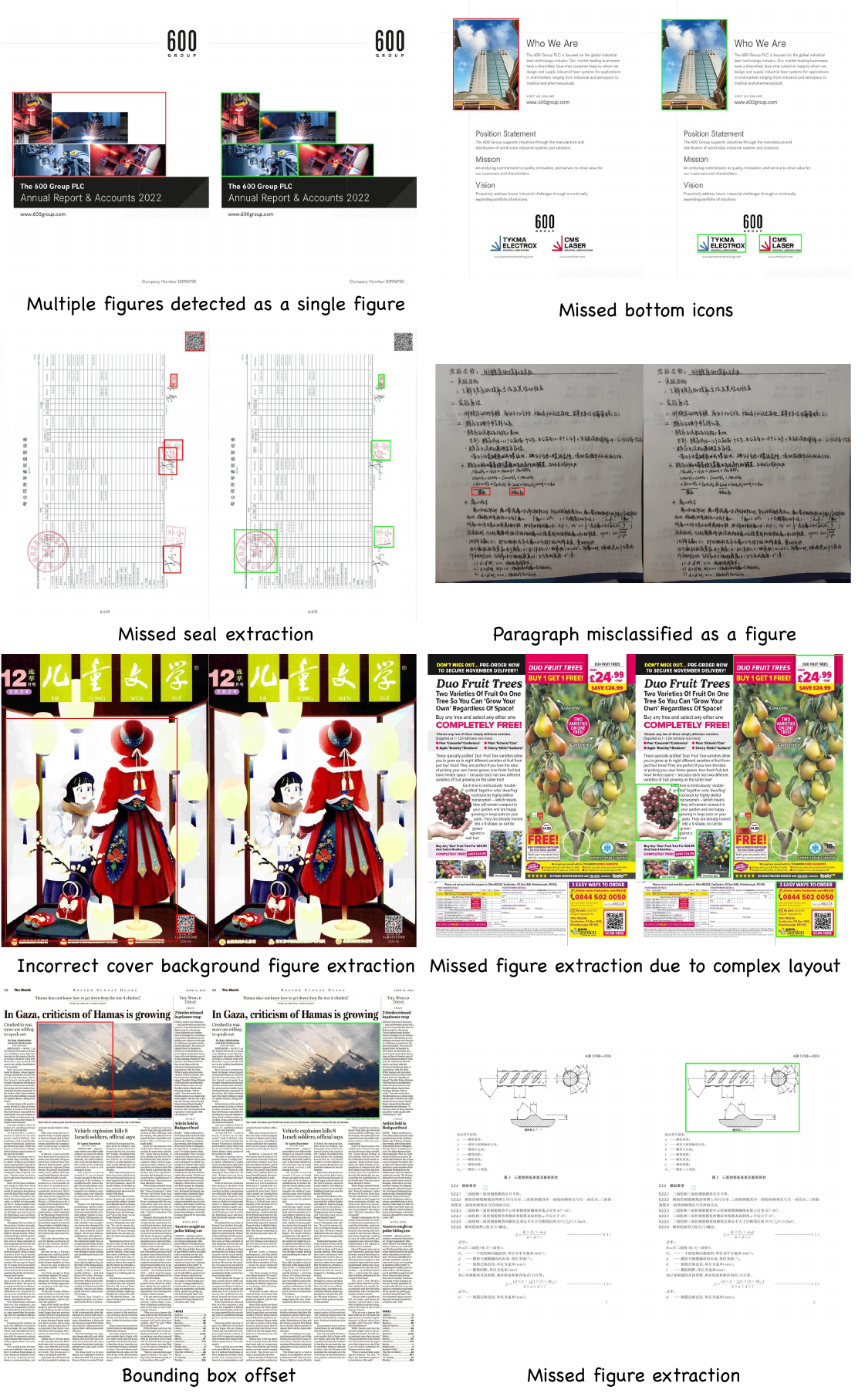}
   \caption{Visualization of figure extraction. The left image shows the model predictions (\textcolor{red}{red boxes}), while the right image presents the ground truth (\textcolor{green}{green boxes}).}\label{fig:supp_figure_extraction}
\end{figure}

\textbf{Figure Extraction.} Fig.~\ref{fig:supp_figure_extraction} presents several visualized results of dots.mocr on the figure extraction task. For each original page, we plot the model predictions and human annotations with rectangular boxes, where the left image shows the predictions and the right image shows the ground truth.
These examples reveal several limitations of current models. First, some figures are missed, indicating weak sensitivity to small, boundary-located, or semantically less salient image regions. Second, some predicted bounding boxes are noticeably shifted and fail to fully cover the figures, suggesting limited accuracy in boundary localization. Third, the model sometimes mistakes cover background images for figures, showing difficulty in distinguishing decorative backgrounds from content-bearing figures. Fourth, performance drops substantially on pages with complex layouts and many visual elements, likely due to stronger visual interference. Finally, for special figure types such as seals, the model is prone to missed detections or incomplete extraction, as these elements are often small, irregular, and overlapping with text or background textures.

\begin{figure}[!h]
    \centering
\includegraphics[width=0.78\linewidth]{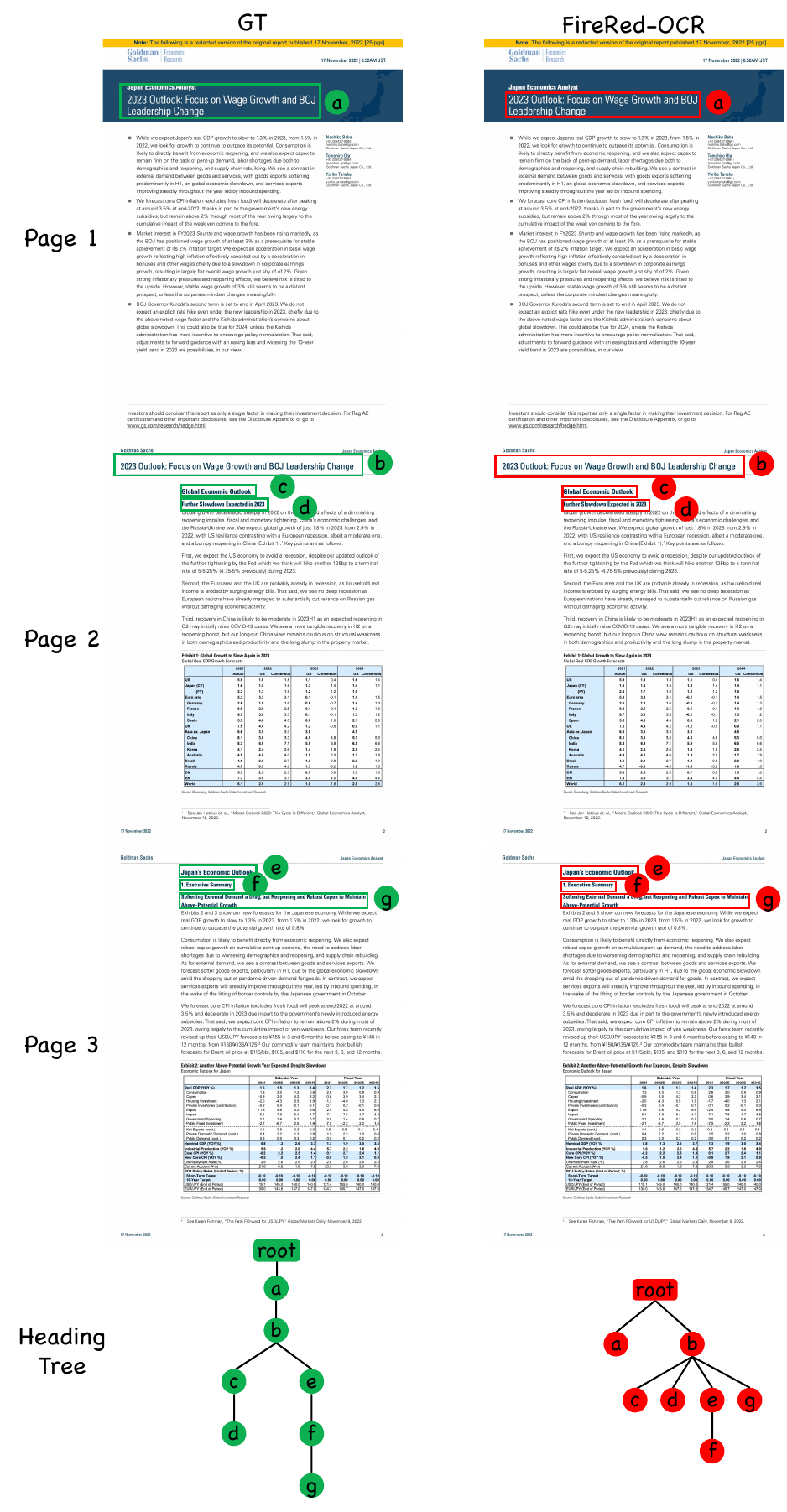}
   \caption{Visualization of heading hierarchy recovery. The \textcolor[RGB]{20,176,83}{left panel} shows the ground-truth heading blocks, while the \textcolor{red}{right panel} shows the predicted heading blocks. The two trees at the bottom represent the ground-truth and predicted heading hierarchies, respectively.
}\label{fig:supp_heading}
\end{figure}

\textbf{Heading hierarchy recovery.} Fig.~\ref{fig:supp_heading} shows a representative example. Although FireRed-OCR is able to predict almost all headings and their corresponding text content, it still makes substantial errors in recovering the hierarchical relationships among headings. Based on both the heading semantics and the formatting cues in the document, it is clear that ``Global Economic Outlook'' and ``Further Slowdown Expected in 2023'' should form a parent-child relation. However, FireRed-OCR incorrectly predicts them as sibling nodes, and similar errors occur repeatedly in this example, eventually resulting in a tree similarity of only 25\%.
This example suggests that current document parsing models remain limited in capturing global heading hierarchies, even when heading detection and content recognition are largely correct. A possible reason is that heading hierarchy recovery requires jointly modeling semantic dependency, visual formatting, and long-range structural relationships across the document, whereas existing models are more focused on local content extraction and element recognition. In future work, it would be valuable to design methods that explicitly incorporate global document structure and hierarchical constraints, so as to improve the robustness of heading hierarchy recovery.

\section{Limitations}\label{limitations}
Our work has several limitations. First, although we adopt a carefully designed annotation pipeline, some annotation errors may still remain in both content and structure annotations. 
In particular, structural annotation of documents inherently involves human judgment, and different annotators may follow slightly different annotation preferences when interpreting the same document. 
As a result, the final annotations may exhibit minor inconsistencies in structural granularity. This issue is difficult to eliminate completely and is common in document parsing benchmarks. 
In future work, we plan to conduct more fine-grained inspection of the dataset, further standardize annotation criteria across annotators, and strengthen the verification process.
Second, for truncated table merging, our current evaluation relies on result-based TEDS. As shown by both the experimental results and qualitative examples, this metric can reflect whether a model performs table merging, but it cannot precisely evaluate whether the merging is carried out in a logically correct way. In future work, we plan to design new evaluation metrics that better capture the correctness of the table merging process.

\section{Copyright and License}\label{copyright}
All data and model weights used in this paper are obtained from publicly accessible sources and are used solely for academic evaluation in document parsing. During data collection, organization, and use, we strictly comply with the copyright policies and licensing agreements of the original sources.

Specifically, part of the data, \ie exam papers and user manuals, are collected from publicly available online resources, including https://ipho.olimpicos.net/, https://www.imo-official.org/problems.aspx, and https://manualzz.com/, \etc These materials are not associated with explicit authorization and are therefore intended solely for academic research purposes and must not be used for any commercial purposes. If any copyright holder has concerns regarding the use of these materials, please contact us through the platform or other designated contact channels, and we will address the issue promptly upon verification.
The dataset as a whole is released under the CC BY 4.0 license. For the portions of the data for which explicit authorization has not been obtained, their use is subject to the CC BY-NC-SA 4.0 license.
For data originating from sources whose original authors or copyright holders cannot be reached, such as https://archive.org/ and https://www.slideshare.net/, we do not directly redistribute the original content. Instead, we provide the corresponding download links so that researchers can access and obtain the data on their own.

\end{document}